\documentclass[10pt,twocolumn,letterpaper]{article}
\usepackage{iccv}
\usepackage{times}
\usepackage{epsfig}
\usepackage{graphicx}
\usepackage{amsmath}
\usepackage{amssymb}
\usepackage[hyphens]{url}
\usepackage{booktabs}
\usepackage{bm}
\usepackage{algorithmicx,algorithm}
\usepackage{multirow}
\usepackage{eqparbox}
\usepackage{color}
\usepackage{float}
\usepackage{makecell}
\usepackage[noend]{algpseudocode}
\usepackage{multirow}
\usepackage{graphbox}
\usepackage{array,booktabs}

\usepackage{comment}
\usepackage{stmaryrd}
\usepackage{breqn}
\DeclareMathOperator*{\argmax}{argmax}

\usepackage[pagebackref=true,breaklinks=true,letterpaper=true,colorlinks,bookmarks=false]{hyperref}
\usepackage[capitalize]{cleveref}
\crefname{section}{Sec.}{Secs.}
\Crefname{section}{Section}{Sections}
\Crefname{table}{Table}{Tables}
\crefname{table}{Tab.}{Tabs.}
\usepackage[accsupp]{axessibility}

\iccvfinalcopy 

\def\iccvPaperID{10622} 

\ificcvfinal\fi

\begin{document}
\title{Collaborative Propagation on Multiple Instance Graphs \\ for 3D Instance Segmentation with Single-point Supervision}

\author{Shichao Dong$^{1,2}$ \quad Ruibo Li$^{1,2}$ \quad Jiacheng Wei$^{2}$ \quad Fayao Liu$^{3}$ \quad Guosheng Lin$^{1,2}$ \thanks{Corresponding author: G.Lin (e-mail:gslin@ntu.edu.sg)}\\
	$^{1}$ S-lab, Nanyang Technological University, Singapore  \\ $^{2}$School of Computer Science and Engineering, Nanyang Technological University, Singapore  \\ $^{3}$  Institute for Infocomm Research, A*STAR, Singapore\\
	{\tt\small \{scdong, gslin\}@ntu.edu.sg} \quad {\tt\small \{ruibo001, jiacheng002\}@e.ntu.edu.sg} \quad {\tt\small fayaoliu@gmail.com} 
}
\maketitle

\ificcvfinal\thispagestyle{empty}\fi

\begin{abstract}
   Instance segmentation on 3D point clouds has been attracting increasing attention due to its wide applications, especially in scene understanding areas. However, most existing methods operate on fully annotated data while manually preparing ground-truth labels at point-level is very cumbersome and labor-intensive. To address this issue, we propose a novel weakly supervised method \textbf{RWSeg} that only requires labeling one object with one point. With these sparse weak labels, we introduce a unified framework with two branches to propagate semantic and instance information respectively to unknown regions using self-attention and a cross-graph random walk method. Specifically, we propose a Cross-graph Competing Random Walks (CRW) algorithm that encourages competition among different instance graphs to resolve ambiguities in closely placed objects, improving instance assignment accuracy. RWSeg generates high-quality instance-level pseudo labels. Experimental results on ScanNet-v2 and S3DIS datasets show that our approach achieves comparable performance with fully-supervised methods and outperforms previous weakly-supervised methods by a substantial margin.
\end{abstract}

\section{Introduction}
\label{sec:intro}
With the rapid development of 3D sensing technology, point cloud based scene understanding has become a popular research topic in recent years. Instance segmentation is one of the most fundamental tasks in this field and has many applications in robotics, autonomous driving, AR/VR, etc. Given a 3D point cloud depicting a scene, this task requires predicting not only a semantic category but also an instance id to differentiate objects at point level. Many deep learning methods have been developed for this task, showing promising results. However, most of these methods operate on point-wise fully annotated data to supervise the network training.

\begin{figure}[t]
    \vspace{-4mm}
	\begin{center}
		\includegraphics[width=1.0\linewidth]{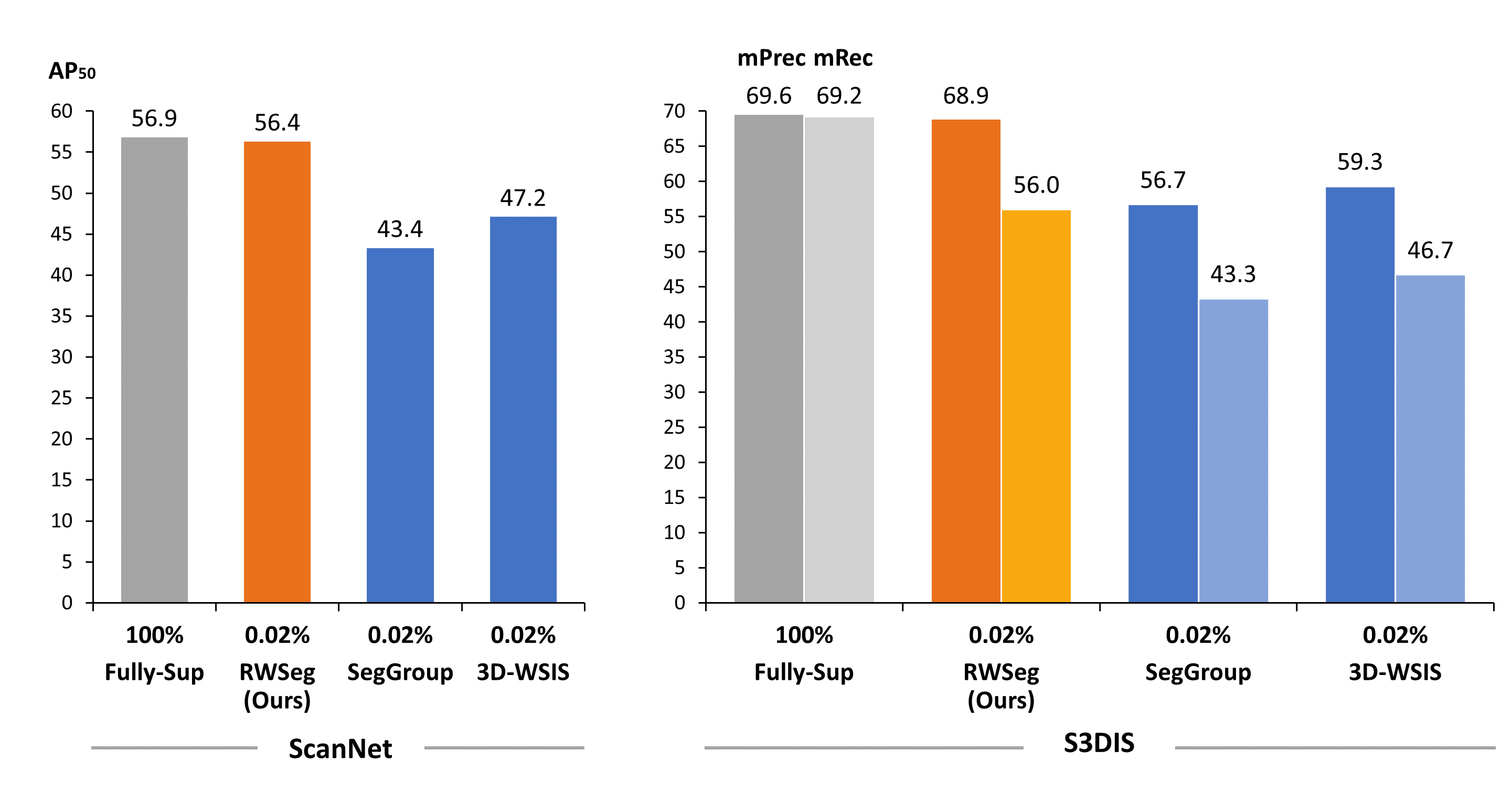}
	\end{center}
	\vspace{-4mm}
	\caption{Comparisons of our approach RWSeg with two recent weakly supervised 3D instance segmentation methods and the fully-supervised baseline on two datasets. Our method achieves better results than other weakly supervised methods with the same amount of weak annotations.}
	\label{fig:performance}
\end{figure}

\begin{figure*}[t]
    \vspace{-8mm}
	\begin{center}
		\includegraphics[width=1.0\linewidth]{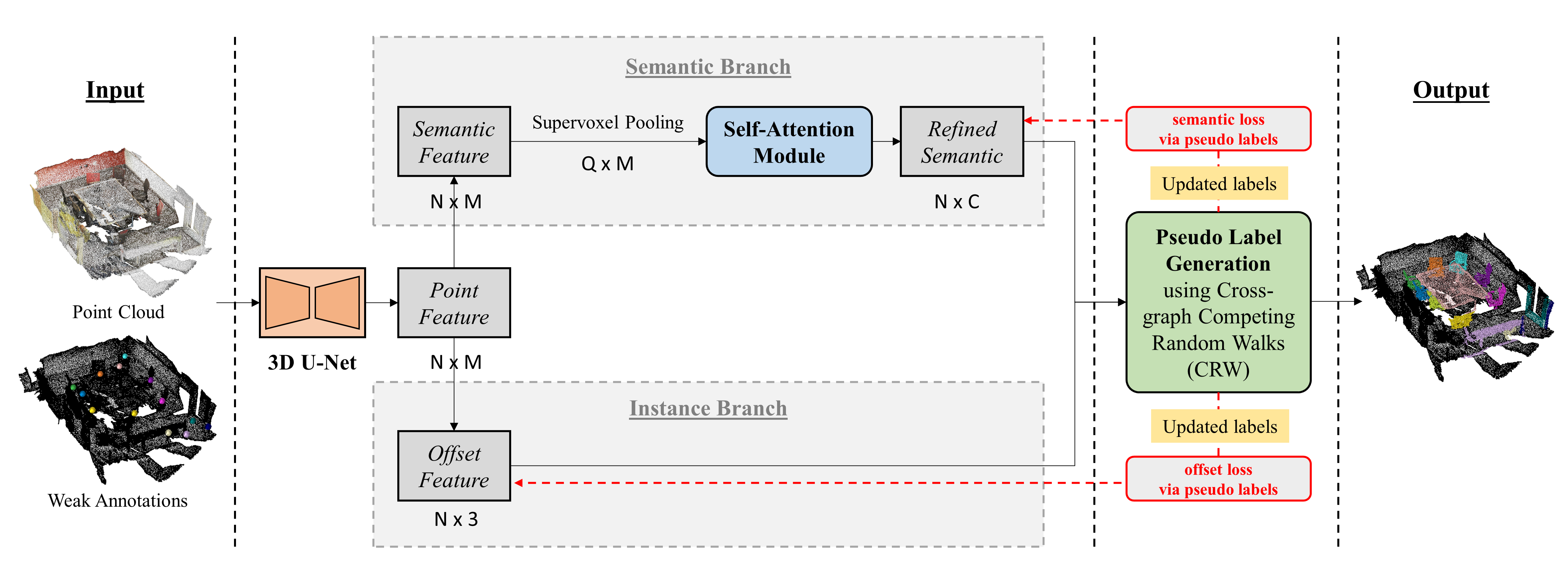}
	\end{center}
	\vspace{-4mm}
	\caption{Pipeline of our proposed weakly supervised method for 3D instance segmentation. The input point cloud is annotated with a single point for each object (enlarged for better visualizations). We use a 3D U-Net backbone based on submanifold sparse convolution \cite{Submanifold} to extract point features. Next, we apply average pooling to the points within the same supervoxel. To facilitate semantic feature propagation, we utilize a self-attention module. Finally, our novel Cross-graph Competing Random Walks (CRW) module leverages the inputs from both branches to generate high-quality pseudo labels for further network training.}
	\label{fig:CGCRW_Pipeline}
\end{figure*}

Manually creating data annotations at point level is very cumbersome and labor-intensive. Although some tools have been adopted to assist, the average time used to annotate one scene is about 22.3 minutes on ScanNet-v2 dataset \cite{dai2017scannet}. To alleviate this issue, several types of weak annotations have been proposed, such as scene-level annotation, subcloud-level annotation \cite{Wei_2020_CVPR}, 2D image based annotation and 3D bounding box annotation \cite{3d_weak_ins_box, chibane2021box2mask}. However, not all weak label types are easy to obtain in practice. In this work, we adopt the annotation method used in SegGroup \cite{tao2020seggroup} and ``One Thing One Click" \cite{otoc_Liu_2021_CVPR}, which only requires annotating a single point for each object. As shown in Figure \ref{fig:CGCRW_Pipeline}, this results in very sparse initial annotations, with less than 0.02\% of total points requiring labeling. According to \cite{tao2020seggroup, otoc_Liu_2021_CVPR},  this annotation method takes less than two minutes per scene, significantly reducing the need for human effort.

Tao et al. \cite{tao2020seggroup} and Tang et al. \cite{tang20223dwsis} have investigated the "One-thing-one-click" approach to address the challenge of weakly supervised 3D instance segmentation. These techniques construct graphs on top of the over-segmentation outcomes and apply Graph Convolution Network (GCN) or inter-superpoint affinity for label propagation. However, these approaches encounter some issues. SegGroup \cite{tao2020seggroup} relies solely on a cross-entropy loss for its semantic prediction with a greedy algorithm for clustering, hence lacking instance-related information. Besides, this method is only designed for the purpose of generating pseudo labels, and therefore requires to utilize these pseudo labels as ground-truth to train a separate network for prediction. 3D-WSIS \cite{tang20223dwsis} utilizes an offset loss and an affinity loss to produce better discriminative features, but their graph is based on the over-segmented point clouds, and the feature of each supervoxel is simply obtained through average pooling of point features and coordinates. The size of supervoxels can vary significantly in their setup, and the initial weak labels can be located at any part of objects, resulting in an unbalanced attraction to neighboring nodes. This may lead to difficulty in identifying precise boundaries, particularly when multiple instances are located close to each other.

In this paper, we propose a novel weakly supervised learning approach, named \textbf{RWSeg}, for 3D point cloud instance segmentation. With only one point annotation per instance, we focus on two key considerations: (1) effective feature propagation is critical for generating high-quality pseudo labels, and (2) leveraging the interactions among instance graphs can be beneficial in finding more accurate instance boundaries and improving the quality of clustering. To address the limitations of previous methods, we are motivated to develop a unified structure for both feature learning and feature propagation.

Convolutional Neural Network (CNN) can extract good local features. However, long-range dependencies can hardly be captured due to its relatively small receptive field. The limitations of CNNs in capturing long-range dependencies are exacerbated in weakly supervised learning scenarios, where only a limited number of certain labels are available to supervise the training process. To this end, we introduce a self-attention module after the 3D CNN backbone, which can effectively propagate long-range information to unknown regions.

For instance pseudo label generation, a customized random walk algorithm on point-level is developed for 3D weakly instance segmentation. The point clouds are first split by their categories, and for each category, multiple instance graphs are built and random walk propagation is performed on each of them. The total energy on each individual graph is identical, based on the assumption that same-class objects tend to have similar sizes. A competing mechanism is designed to perform collaborative propagation on multiple instance graphs. To sum up, the key contributions of our work are as follows:

\begin{itemize}
    \item We design a unified framework for weakly supervised 3D instance segmentation. To enhance the feature propagation, we introduce a self-attention module to capture long-range dependencies.

    \item We propose a novel algorithm to perform collaborative propagation on multiple instance graphs to generate high-quality instance pseudo labels. The designed competing mechanism helps to resolve ambiguous cases in 3D instance segmentation task.
    
    \item With significantly fewer annotations, our method bridges the gap between weakly supervised learning and fully supervised learning in 3D instance segmentation.
\end{itemize}

\section{Related Work}
\label{sec:rltd}
\paragraph{Fully supervised 3D segmentation}
To effectively process unstructured and unordered 3D data , current feature learning methods can be broadly categorized into two types: point-based methods \cite{PointCNN, PointNet, PointNet++, KPConv, PointConv, PointWeb, 9552005, Guo_2021} and voxel-based methods \cite{multiview2, Submanifold, MVPNet, multiview}. Voxel-based approaches involve transforming data into 3D volumetric grids, whereas point-based methods operate directly on the individual points. Instance segmentation on point clouds can be seen as a joint task of segmentation and localization. Proposal-based methods \cite{SGPN, 3DSIS, 3DBoNet, 3dmpa} detect object boundaries explicitly and then perform binary mask segmentation as the final output. On the other hand, proposal-free methods \cite{SGPN, MASC, ASIS, JSIS3D, MTML, pointgroup, occuseg, Chen_2021_ICCV, Liang_2021_ICCV, dong2022learning} directly regress instance centroids without performing the detection task. Jiang et al. \cite{pointgroup} utilized a submanifold sparse convolution \cite{Submanifold} based 3D U-net and proposed to use a breadth-first search clustering algorithm on dual coordinate sets.

\paragraph{Weakly supervised segmentation}
Numerous weakly supervised methods have been proposed for image segmentation \cite{ahn2018learning,huang2018weakly,pinheiro2015image,papandreou2015weakly,ahn2019weakly,zhou2018weakly,arun2020weakly}. Wei et al. \cite{wei2020multi} proposed the first weakly supervised approach for point cloud semantic segmentation, utilizing Class Activation Map (CAM) to generate point-level pseudo labels with subcloud-level annotations. Several subsequent works \cite{xu2020weakly, wang2020weakly, otoc_Liu_2021_CVPR, dong2023weakly} also addressed weakly segmentation on point clouds with lesser supervision. There have been limited attempts to solve 3D weakly supervised instance segmentation. Hou et al. \cite{hou2021exploring} designed a pre-training method to assist prediction, while Tao et al. \cite{tao2020seggroup} proposed Seggroup with graph convolution network (GCN) for instance label propagation. However, Seggroup lacks the ability to learn discriminative features for separating instances. Tang et al. \cite{tang20223dwsis} proposed to learn discriminative features and use inter-superpoint affinity for label propagation. However, their method did not fully utilize all the spatial information and may affect their performance on ambiguous cases. Liao et al. \cite{3d_weak_ins_box} and J. Chibane et al. \cite{chibane2021box2mask} proposed using 3D bounding boxes as supervision. However, box annotation provides much richer information than clicking one point per instance, and most non-overlapped objects can already be defined by 3D bounding box. This may lessen the significance of their work.

\section{Method}
\label{sec:mthd}
In this section, we first introduce our data annotation setting for point cloud instance segmentation in Section \ref{sec:anno}. Then, Section \ref{sec:stgy} describes our training strategy. Lastly, Section \ref{sec:ntwk} and \ref{sec:psd} present our approach in detail, including network architecture, semantic branch, instance branch and proposed pseudo label generation algorithm.

\subsection{Weak Annotation}
\label{sec:anno}
Following SegGroup \cite{tao2020seggroup}, we adopt the annotation setting of one point per object, as shown in Figure \ref{fig:CGCRW_Pipeline}. To create initial pseudo labels, we spread the labels from annotated points to nearby points within the same supervoxel segment. These segments are generated by unsupervised over-segmentation method~\cite{felzenszwalb2004efficient} based on the surface normals of points. Points within the same segment have high internal consistency, which are used as the initial ground-truth to supervise the network training.

\begin{figure}[t]
	\begin{center}
		\includegraphics[width=1.0\linewidth]{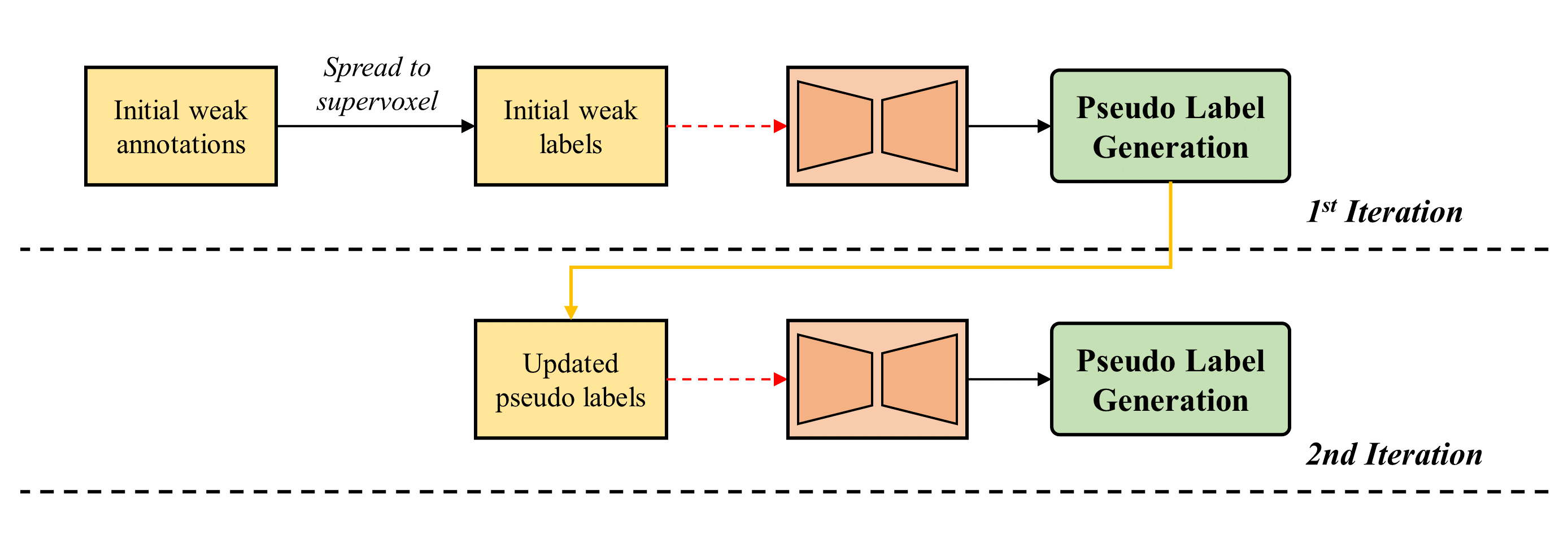}
	\end{center}
	\vspace{-6mm}
	\caption{Learning cycle of our proposed weakly supervised method for 3D Instance Segmentation.}
	\vspace{-4mm}
	\label{fig:RWSeg_cycle}
\end{figure}

\subsection{Learning Strategy}
\label{sec:stgy}
As shown in Figure \ref{fig:RWSeg_cycle}, the network training of our method consists of two stages. The first stage is supervised by the initial weak labels. Afterward, predictions with high confidence from our pseudo label generation algorithm are further updated as new ground-truth labels for next stage training. With this learning strategy, the quality of learned features can be consistently improved

\subsection{Network Architecture}
\label{sec:ntwk}
Our network takes point cloud $\bm P \in \mathbb{R}^{N \times 3}$ as input where $N$ is the number of points in $\bm P$. It uses a shared U-Net backbone and two separate branches for point-level semantic feature learning and instance centroid regression. In the semantic branch, a self-attention based module is used to further enhance semantic features, especially for those regions without supervision. Following that, our proposed \textbf{Cross-Graph Competing Random Walks (CRW)} algorithm leverages learned features and existing ground-truth weak labels to generate instance-level pseudo labels. With refined weak labels, the network can be further trained to produce better features. All proposed modules are within the unified framework, as shown in Figure \ref{fig:CGCRW_Pipeline}.

\paragraph{Semantic segmentation branch}
The submanifold sparse convolution \cite{Submanifold} based backbone network can extract point-wise features with good local information capturing. However, to enhance the network's ability to capture long-range feature dependencies and extend its receptive field, we propose incorporating a self-attention module to further refine the semantic features. In order to reduce the computational complexity of self-attention and ensure local geometric consistency, we utilize a supervoxel generation method \cite{Lin2018Supervoxel}. Specifically, for each supervoxel set, we apply average pooling to both point coordinates and semantic features. Following \cite{NIPS2017_3f5ee243, zhao2021pointtransformer}, we build a self-attention layer across all the supervoxels and then interpolate the output to the original size of the input point cloud. During training, we use a conventional cross-entropy loss $H_{CE}$ with incomplete labels to supervise the process. The structure diagram and formulas of the self-attention module are provided in the supplementary.

\paragraph{Instance centroid offset branch} 
Parallel to the semantic branch, we apply a 2-layers MLP upon point features to predict point-wise centroid shift vector $d_i \in \mathbb{R}^{3}$. The instance
centroid $\widehat{q}$ is defined as the mean coordinates of all points with the same instance label. Following \cite{pointgroup}, We use an $L_{1}$ regression loss and a cosine similarity based direction loss to train the offset prediction. We only consider foreground points with weak labels or pseudo weak labels for supervision. With initial weak labels, real centroids of instances can hardly be inferred. However, we found it is still beneficial to apply offset loss, as it can help to slightly shift points towards inner part of objects. 

The final joint loss function can be written as
\begin{equation}
	L_{joint} = L_{sem} + L_{offset}.
\end{equation}

\subsection{Pseudo label Generation}
\label{sec:psd}
After training with the initial weak labels, we now have a network that can make semantic prediction and offset prediction, which can be further utilized to generate pseudo instance labels. However, due to the limited supervision used during model training, the quality of the prediction may not be very accurate at the first iteration. To address this issue, we propose a random walk-based algorithm to generate reliable pseudo labels for unlabelled points. 

\begin{figure}[t]
	\begin{center}
		\includegraphics[width=1.0\linewidth]{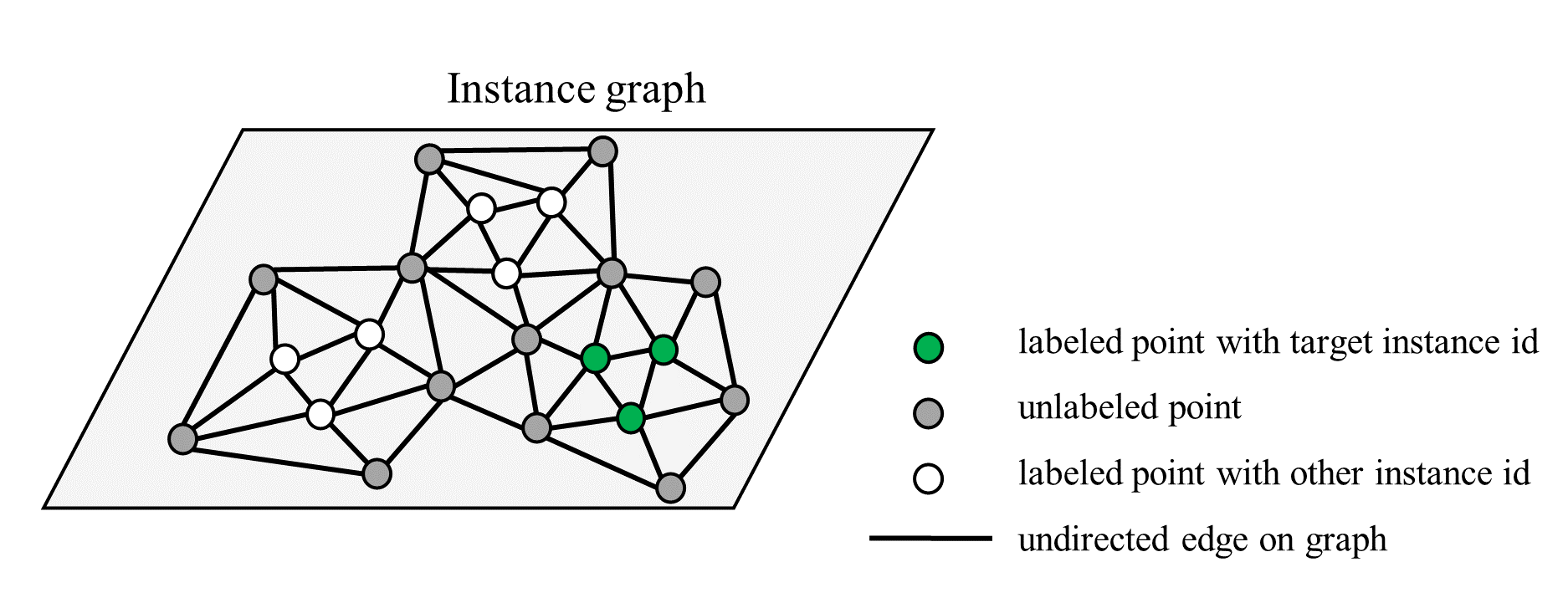}
	\end{center}
	\vspace{-4mm}
	\caption{Illustration of a single instance graph. The nodes on graph are connected by undirected edges. The edge weights are determined by the transition matrix $\bm A$ in Eq. (\ref{eq:A_norm}). The initial node values are determined by the vector ${\bm b}_{0}$ in Eq. (\ref{eq:b_init}) For this example, the initial values of the three green nodes are $1/3$.} 
	\label{fig:single_graph}
\end{figure}

\begin{figure*}[t]
    \vspace{-2mm}
	\begin{center}
		\includegraphics[width=1.0\linewidth]{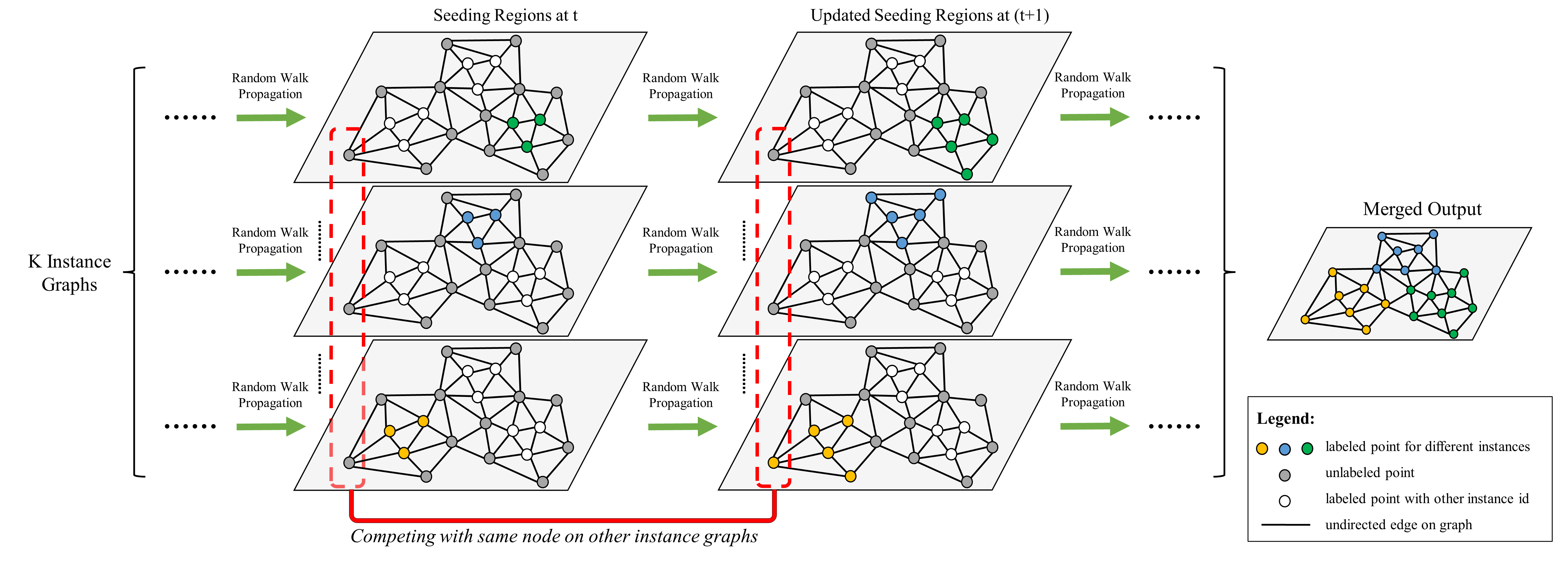}
	\end{center}
	\vspace{-6mm}
	\caption{Illustration on Cross-graph Competing Random Walks (CRW). Our algorithm takes a group of points from the same semantic category as input and constructs $K$ graphs according to the number of instances. Proposed method enables the interactions among the same positioned nodes on $K$ instance graphs. Point score can be suppressed or enhanced after cross-graph competition, thereby affecting the following seeding points update strategy. High score points enjoy the priority to be grouped first. After performing several iterations, instance graphs are merged to generate the final output prediction.}
	\label{fig:CGCRW}
\end{figure*}

 In this section, we first describe how we construct an individual graph in Figure \ref{fig:single_graph} and then present the details of cross-graph competing mechanism and the clustering algorithm in Figure \ref{fig:CGCRW}. The core idea of our algorithm is to enable interactions among instance graphs and gradually updates seeding points until reaching a signal equilibrium state.

\vspace{-2mm}
\paragraph{Building graph on the point cloud} 
According to the semantic predictions from the semantic branch $\mathbf{S} = \{s_1, s_2, ..., s_N\}\in \mathbb{R}^{N}$ on point clouds $\bm P$, we treat each foreground semantic category as a target group. For each group, we build $K$ fully connected and undirected instance graphs, with $K$ being the number of instances. As shown in Figure \ref{fig:single_graph}, the nodes of each graph are points from all $K$ instances. Each node in each graph is associated with an initial binary label (score), as detailed in the paragraph below. The $K$ instance graphs have the same nodes and edges, with the only difference being they have different initial graph node score vectors.

For the $l$-th instance graph, its initial graph node score vector $\bm b^l_0$ is defined by its binary instance label mask $\bm m^l$, with the $i$-th element $m^l_i=1$ if the $i$-th node (point) has an instance label of $l$. The $i$-th element of $\bm b^l_0$ is:

\begin{equation}\label{eq:b_init}
	{b^{l(i)}_{0}} =
	\begin{cases}
		 \frac{1}{\sum_{j=0}^{n}{m^l_j}} & {m^l_i} = 1 \\
		 $0$ & otherwise,
	\end{cases}
\end{equation}

where $n$ is the number of nodes in the graph. The process of normalizing seeding points' initial scores involves dividing them by the total number of nodes corresponding to the same instance id. This normalization aims to achieve equitable allocation of the initial potential among instance graphs, thereby preventing any undue advantage for instances having a larger number of positive weak labels.

The random walk operation on each graph can be modeled with an~$n \times n$ transition matrix~$\bm A$. $\bm A_{ij} \in [0, 1]$ denotes the transition probability between $i$-th and $j$-th nodes, with a higher value indicating a higher transition probability.

To build transition matrix~$\bm A$, we first consider a pairwise kernel function to derive a symmetric affinity matrix $\bm W$, which helps to enhance local smoothness. For each edge connecting the $i$-th and $j$-th nodes, we define its weight as:

\begin{equation}\label{eq:W2}
	{\bm W}_{ij} = {\rm \exp}  (- \frac{\| (x_i + d_i) - (x_j + d_j) \|^2}{2\sigma^2}),
\end{equation}

where $\sigma$ is a hyperparameter, $x_i$ and $x_j$ are point coordinates. We use the predicted offset vector $d$ from the instance branch to shift points from their original coordinates toward their instance centers. Node pairs with small euclidean distances in the 3D space tend to have high similarities. 

Next, we formulate the transition matrix $\bm A$ by the following rules. Specifically, we assign a weight of zero to the edges connecting nodes belonging to different instance labels, in order to restrict the direct interaction between them.

\begin{equation}\label{eq:A}
	\bm A_{ij} = 
	\begin{cases}
		 $0$ & \widehat{\bm L}_{i} \neq \widehat{\bm L}_{j}\\
		\bm W_{ij} & otherwise
	\end{cases},
\end{equation}

where $\widehat{\bm L}_{i}$ and $\widehat{\bm L}_{j}$ are the instance labels of two nodes. Lastly, transition matrix $\bm A$ needs to be normalized:

\begin{equation}\label{eq:A_norm}
	{\bm A}_{ij} = \frac{{\bm A}_{ij}}{\sum_{j \in n}{\bm A}_{ij}}.
\end{equation}

This transition matrix $\bm A$ is shared among each group of instance graphs. 

Random walk algorithm is performed by repeatedly adjusting node vector $\bm b$ via the transition matrix $\bm A$. At $t$-th iteration, the adjustment can be expressed as

\begin{equation}\label{eq:rw_1}
    {\bm b}^l_{t+1} = \alpha \bm A \bm b^l_{t} + (1-\alpha)\bm b^l_{0}, 
\end{equation}

where $\alpha \in [0, 1]$ is a blending coefficient between propagated scores and the initial scores.

When repeatably applying unlimited random steps on a graph, it will reach equilibrium. The final steady-state of random walk algorithm can be written as
\begin{equation}
    {\bm b^l_{(\infty)}}   =  (1 - \alpha) ({\bm I} -\alpha {\bm A})^{-1}  \bm b^l_0.
\end{equation}

\begin{algorithm}[hbt] 
    \caption{\small Cross-graph Competing Random Walks (CRW)}
	\hspace*{0.02in} {\bf Input:} coordinates $\mathbf{X} = \{x_1, x_2, ..., x_N\}\in \mathbb{R}^{N \times 3}$; number of instances per category $K = \{k_1, k_2, ..., k_{s}\}$ ($s$ is the total number of valid classes); hyperparameter $\alpha, \theta$; max iteration number ${t_1}_{max}$ and ${t_2}_{max}$; instance weak labels $\widehat{\bm L}$; semantic prediction $\bm S$; offset prediction $\bm D$ \\
	\hspace*{0.02in}{\bf Output:} 
	Instance pseudo label prediction $\bm P$ \\
	\vspace*{-3mm}
	\begin{algorithmic}[1]
	    \For {$id$ in foreground semantic IDs}
	        \For {$\bm S \in id$}
    	    \State build $K$ instance graphs ;
    	    \State construct affinity matrix $\bm W$ via Eq. (\ref{eq:W2});
    	    \State construct transition matrix $\bm A$ via Eq. (\ref{eq:A}); 
    	    \State normalize transition matrix $\bm A$ via Eq. (\ref{eq:A_norm});
    	    \For {$l = 1$ to $K$}
    	    \State initialize graph node vector via Eq. (\ref{eq:b_init});
    	    \EndFor
    	    \While{ $t_1 \leq {t_1}_{max}$ }
    	    \For {$l = 1$ to $K$}
    	    \State propagate one step via Eq. (\ref{eq:rw_1});
    	    \EndFor
    	    \State $t_1 \gets t_1 + 1$
    	    \EndWhile
    	    \While{ $t_2 \leq {t_2}_{max}$ }
    	    \State adjust node vectors via Eq. (\ref{eq:softmax})
    	    \For {$l = 1$ to $K$}
    	    \State reinitialize vector via Eq. (\ref{eq:b_init});
    	    \State update top $\theta$ as new seeding points
    	    \State propagate one step via Eq. (\ref{eq:rw_1});
    	    \EndFor
    	    \State $t_2 \gets t_2 + 1$
    	    \EndWhile
    	    \State $p_{i} \gets \argmax{(\bm b^{(i)})}$ 
    	    \State $p_{i} \gets \widehat{\bm L}_{j}$ if under the same mask
    	    \EndFor
	    \EndFor
		\State \Return $\mathbf{P}$
	\end{algorithmic}
\end{algorithm}

\begin{figure*}[thb]
	\begin{center}
		\includegraphics[width=1.0\linewidth]{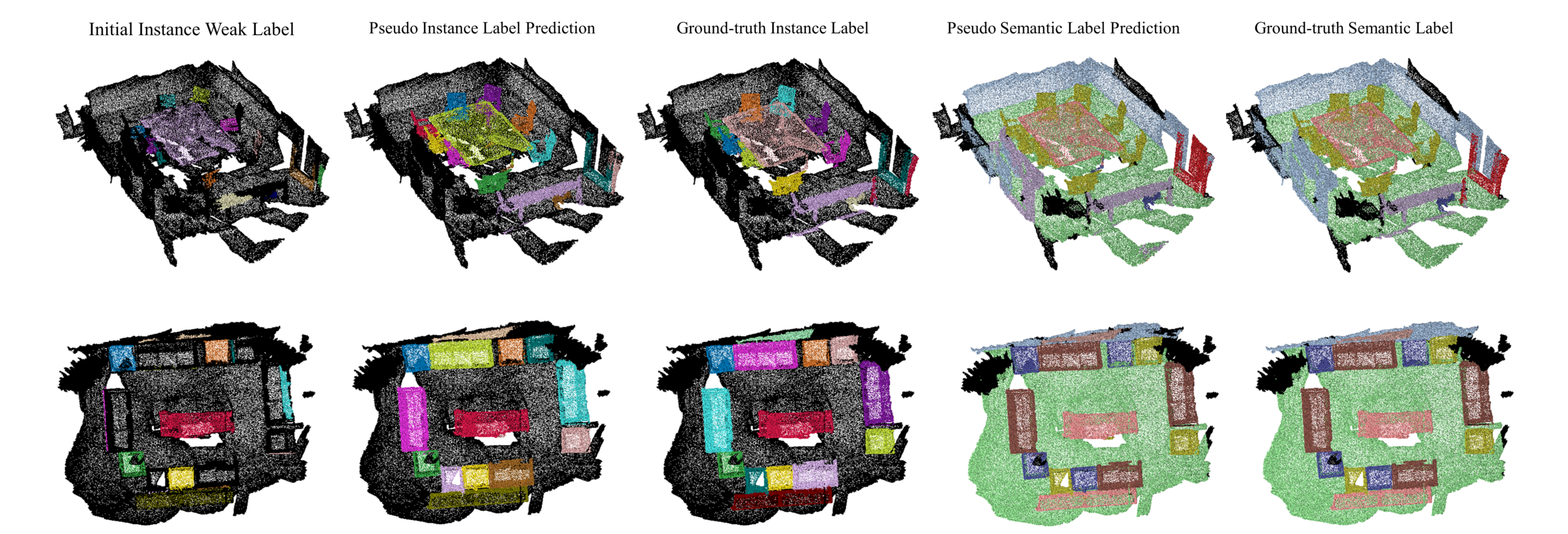}
	\end{center}
	\caption{The qualitative visualization results of generated pseudo labels on ScanNet-v2 dataset\cite{dai2017scannet}}
	\label{fig:scannet_result}
\end{figure*}

\begin{figure*}[thb]
	\begin{center}
		\includegraphics[width=1.0\linewidth]{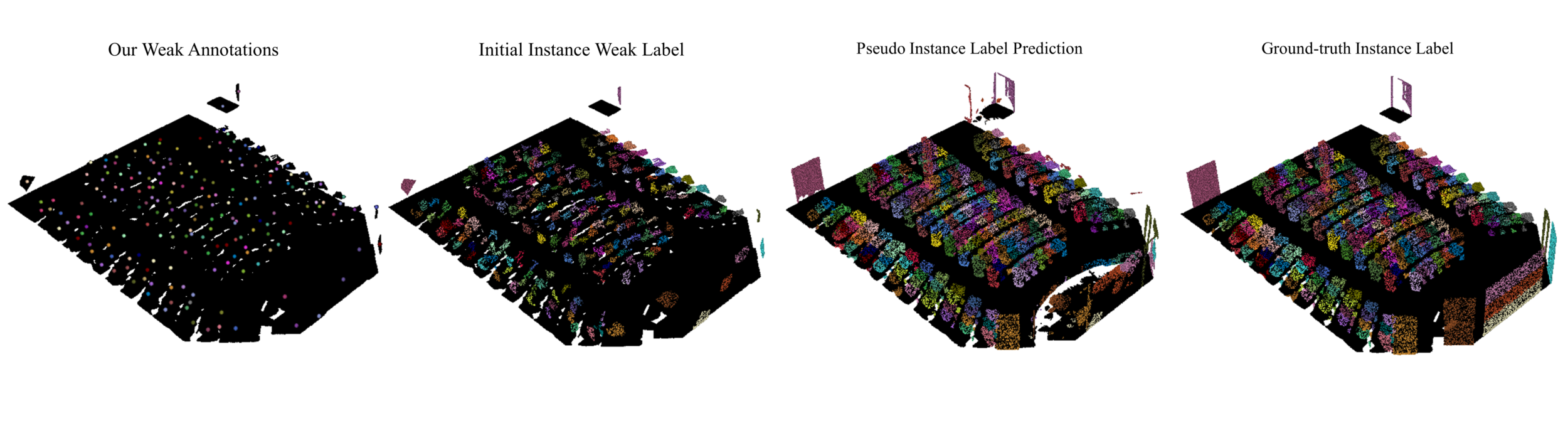}
	\end{center}
	\vspace{-12mm}
	\caption{The qualitative visualization results of generated pseudo labels on S3DIS dataset \cite{S3DIS}}
	\label{fig:s3dis_result}
\end{figure*}

\paragraph{Cross-graph Competing Random Walks (CRW)}
On top of the random walk algorithm, we design a mechanism to encourage competitive interactions among instance graphs, in Figure \ref{fig:CGCRW}. Intuitively, the idea is to suppress a point's activation score in the current graph if its scores in other instance graphs are relatively high. However, the level of repulsive effect needs to be well controlled. Otherwise, too strong repulsive effects are likely to distort the results from the random walk.

Based on the random walk results, we apply a softmax function to every node score to adjust the probability distribution over the $K$ instance graphs. Elements in the score vector are re-scaled to the range of $[0,1]$, and the score values of the same positioned nodes on $K$ instance graphs are summed to $1$.

\begin{equation}\label{eq:softmax}
    b^{l(i)} = \frac{exp(b^{l(i)})}{\sum_{j=1}^K{exp(b^{j(i)}})},
\end{equation}

where $b^{l(i)}$ denotes the score of the $i$-th node on $l$-th graph. In this simple manner, we bring repulsive interaction among instance graphs. A point that receives less competition from other instance graphs will be adjusted to a relatively higher score, and vice versa.

Then, for each instance, we pick a fixed percentage $\theta$ (i.e. $50\%$) of newly predicted pseudo labels with high confidence to be updated as seeding points for the next iteration. The selection is based on the sorted node scores. Only unlabelled points can be considered as new seeding points.
Our approach gradually groups relatively confident points into seeds and performs a random walk step at each iteration.

\begin{table*}[tb]
	\footnotesize
	\begin{center}
		\begin{tabular}{l|c|l|l}
			\toprule
			\textbf{Semantic mIoU} & Label& wall~\,floor~\,cab~~\,bed~~\,chair~\,sofa~\,table~\,door~\,wind~\,bkshf~~pic~~~cntr~~\,desk~~curt~\;fridg~\,showr~\,toil~~\,sink~~bath~\,ofurn & ~avg\\
			\midrule
			MPRM~\cite{wei2020multi} & Scene & 47.3~\,41.1~\,10.4~\,43.2~~25.2~~43.1~\;21.9~\;\;9.8\;\;\,12.3~~\,45.0~~\,\;9.0\;\;\,13.9~~21.1~~40.9~~\;1.8\;\;\,\,29.4~~\,14.3~\,\;9.2~\;\,39.9~~10.0&~24.4\\
			MPRM~\cite{wei2020multi} & Subcloud & 58.0~\,57.3~\,33.2~\,71.8~~50.4~~69.8~\;47.9~\;42.1~~44.9~~\,73.8~~\,28.0~~21.5~~49.5~~72.0~~38.8~~\,44.1~~\,42.4~\,20.0~~48.7~~34.4&~47.4\\
			\midrule
			SegGroup \cite{tao2020seggroup} & 0.02\%  & 71.0~\,82.5~\,63.0~\,52.3~~72.7~~61.2~\;65.1~\;66.7~~55.9~~\,46.3~~\,42.7~~50.9~~50.6~~67.9~~67.3~~\,70.3~~\,70.7~\,53.1~~54.5~~63.7&~61.4\\
			\midrule
			\midrule
		    \textbf{RWSeg (Ours)}  & 0.02\%  & \textbf{88.8}~\,\textbf{94.4}~\,\textbf{80.2}~\,\textbf{82.4~~}\textbf{85.9}~~\textbf{91.2}~\;\textbf{76.5}~\;\textbf{76.6}~~\textbf{78.2}~~\,\textbf{87.5}~~\,\textbf{66.3}~~\textbf{64.1}~~\textbf{67.7}~~\textbf{85.6}~~\textbf{86.9}~~\,\textbf{88.9}~~\,\textbf{92.4}~\,\textbf{71.5}~~\textbf{91.7}~~\textbf{75.3}&~\textbf{81.6}\\
			\bottomrule
		\end{tabular}
	\end{center}
	\vspace{-2mm}
	\caption{Pseudo label quality of semantic segmentation (category-level) on ScanNet-2 \cite{dai2017scannet} training set. }
	\label{sem_iou_train}
\end{table*}

\begin{table*}[tb]
	\footnotesize
	\begin{center}
		\begin{tabular}{l|c|l|l}
			\toprule
			\textbf{Instance AP} & Metric& cab~~\,bed~~\,chair~\,sofa~\,table~\,door~\,wind~\,bkshf~~pic~~~cntr~~\,desk~~curt~\;fridg~\,showr~\,toil~~\,sink~~bath~\,ofurn & ~avg\\
			\midrule
			\midrule
			\multirow{3}{*}{\textbf{RWSeg (Ours)}} & AP & 59.0~\,65.9~~70.3~~82.1~\;59.3~\;38.2~~54.0~~\,68.0~~\,54.7~~35.6~~35.8~~48.0~~73.9~~\,80.6~~\,88.4~\,44.6~~85.4~~54.3&~\textbf{61.0}\\
			& AP{\tiny 50} & 85.7~\,94.2~~93.8~~90.3~\;87.1~\;60.8~~77.1~~\,84.5~~\,81.6~~79.9~~74.1~~69.7~~92.1~~\,92.4~~\,97.8~\,82.3~~97.4~~77.6&~\textbf{84.4}\\
			& AP{\tiny 25} & 96.4~\,99.0~~98.3~~95.7~\;95.2~\;87.0~~91.6~~\,91.5~~\,92.8~~96.3~~93.9~~87.5~~99.2~~\,97.4~~\,99.3~\,95.8~100.0~92.8&~\textbf{95.0}\\
			\bottomrule
		\end{tabular}
	\vspace{-2mm}
	\end{center}
	\caption{Pseudo label quality of instance segmentation on ScanNet-2 \cite{dai2017scannet} training set. }
	\label{ins_ap_train}
\end{table*}

\section{Experiments}
\label{sec:exp}

\paragraph{Datasets}
In this section, we show our experimental results on two public datasets: ScanNet-v2 \cite{dai2017scannet} and S3DIS \cite{S3DIS} to show the effectiveness of our proposed method. ScanNet-v2 dataset \cite{dai2017scannet} is a popular 3D indoor dataset containing 2.5 million RGB-D views in 1513 real-world scenes, covering 20 semantic categories. The evaluation metrics of 3D instance segmentation are mean average precisions at different overlap percentages, i.e., mAP@0.25, mAP@0.5 and mAP respectively. S3DIS dataset \cite{S3DIS} has 272 scenes under six large-scale indoor areas. Unlike ScanNet \cite{dai2017scannet}, all 13 classes including background are annotated as instances and require prediction. We use the mean precision (mPre) and mean recall (mRec) with an IoU threshold of 0.5 as the evaluation metric. 

\vspace{-4mm}
\paragraph{Implementation details}
We set the voxel size as $2 cm$ for submanifold sparse convolution \cite{Submanifold} based backbone, following \cite{pointgroup}. Our network is trained on a single GPU card. For each stage of training, the backbone network and self-attention module are trained sequentially, with a batch size of 4 and 2 respectively. We set $\gamma$ and $\delta$ in the self-attention module as two-layer MLPs with the hidden dimension of 64 and 32 respectively. For CRW algorithm, we set hyerparameters $\alpha = 0.2, {t_1}_{max} = 1, {t_2}_{max} = 5$ and $\theta$ as 50\%. Due to GPU memory limit, we subsample the input point cloud to CRW if the point number is above 25k. Last output remains at original resolution. For network training, we use Adam solver for optimization with an initial learning rate of 0.001. 

\paragraph{Pseudo label evaluation}
As shown in Table \ref{sem_iou_train} and Table \ref{ins_ap_train}, we present the quality of our generated pseudo labels based. Reported final pseudo labels are created after two stages of network training. Our network is trained only on the training set of ScanNet-v2 \cite{dai2017scannet} with 1201 scenes, no extra data is needed. In Table \ref{sem_iou_train}, the semantic quality of our pseudo labels largely outperforms previous methods by at least $20.2\%$. Besides, we also report the instance quality of pseudo labels in Table \ref{ins_ap_train}. However, no available data from other methods can be used for comparison at present. Our qualitative pseudo labels can be used by any fully supervised method to resolve their annotation cost issue.

\paragraph{Prediction evaluation}
Different from weakly supervised methods like SegGroup \cite{tao2020seggroup} that require training another a new network for prediction, we can directly adopt other methods on the same network for prediction without retraining. Here we employ a Breadth-First Search (BFS) clustering algorithm from PointGroup \cite{pointgroup} to our network. In Table \ref{ins_val}, we compare the prediction results with fully supervised PointGroup \cite{pointgroup} and other weakly supervised methods on ScanNet-v2 \cite{dai2017scannet} validation set. 

Our method significantly outperforms SegGroup \cite{tao2020seggroup} and 3D-WSIS \cite{tang20223dwsis} over all evaluation metrics, generally \textbf{ by an absolute margin of around 10 points}. Remarkably, with only $0.02\%$ of annotated points, we \textbf{achieve comparable results with fully supervised method \cite{pointgroup}}. We also report the instance segmentation results on ScanNet-v2 \cite{dai2017scannet} test set in Table \ref{ins_test}. Our method again performs significantly better than other weakly supervised methods which use the same amount of annotations. For S3DIS \cite{S3DIS} dataset, we report Area 5 and 6-fold cross validation results in Table \ref{ins_s3dis}.

\begin{table}[t]
	\footnotesize
	\begin{center}
		\begin{tabular}{l|c|ccc}
			\toprule
			Method  & Supervision  & ~AP & \hspace{-3pt}AP{\tiny 50}\hspace{-3pt} & AP{\tiny 25}\\
			\midrule
			Full Supervision: &&&&\\
			PointGroup~\cite{pointgroup} & 100\% & 34.8 & \hspace{-3pt}56.9\hspace{-3pt} & 71.3\\
			\midrule
			Init+Act. Point Supervision: &&&&\\
			CSC-20 (PointGroup)~\cite{hou2021exploring} & 20 pts/scene & - & \hspace{-3pt}27.2\hspace{-3pt} & - \\
			CSC-50 (PointGroup)~\cite{hou2021exploring} & 50 pts/scene & - & \hspace{-3pt}35.7\hspace{-3pt} & - \\
			\midrule
			SPIB \cite{3d_weak_ins_box} & 100\% Box & - & \hspace{-3pt}38.6\hspace{-3pt} & 61.4 \\
            Box2Mask \cite{chibane2021box2mask} & 100\% Box & - & \hspace{-3pt}59.7\hspace{-3pt} & 71.8 \\
			\midrule
            TWIST \cite{TWIST9879061} & 1\% & 9.6 & \hspace{-3pt}17.1\hspace{-3pt} & 26.2 \\
			TWIST \cite{TWIST9879061} & 5\% & 27.0 & \hspace{-3pt}44.1\hspace{-3pt} & 56.2 \\
			TWIST \cite{TWIST9879061} & 10\% & 30.6 & \hspace{-3pt}49.7\hspace{-3pt} & 63.0 \\
			TWIST \cite{TWIST9879061} & 20\% & 32.8 & \hspace{-3pt}52.9\hspace{-3pt} & 66.8 \\
			\midrule
			One Obj One Pt Supervision: &&&&\\
			SegGroup (PointGroup) \cite{tao2020seggroup} &0.02\% & ~23.4 & \hspace{-3pt}43.4\hspace{-3pt} & 62.9 \\
            3D-WSIS \cite{tang20223dwsis} & 0.02\% & 28.1 & \hspace{-3pt}47.2\hspace{-3pt} & 67.5 \\
			\midrule
			\midrule
			\textbf{RWSeg (Ours)} &0.02\% & ~34.7 & \hspace{-3pt}56.4\hspace{-3pt} & 71.2 \\
			\bottomrule
		\end{tabular}
	\end{center}
	\vspace{-2mm}
	\caption{Instance segmentation results on ScanNet-v2 \cite{dai2017scannet} validation set. Methods marked with brackets represents using generated pseudo labels to train another fully-supervised method.}
	\label{ins_val}
\end{table}

\begin{table}[t]
	\footnotesize
	\begin{center}
		\begin{tabular}{l|c|ccc}
			\toprule
			Method & Supervision & ~AP & AP{\tiny 50} & ~AP{\tiny 25}~\\
			\midrule
			Full Supervision:~~ &&&&\\
            SoftGroup \cite{vu2022softgroup} & 100\% & ~50.4 & 76.1 & 86.5~\\
            HAIS \cite{Chen_HAIS_2021_ICCV} & 100\% & ~45.7 & 69.9 & 80.3~\\
            SSTNet \cite{SSTNet_liang2021instance} & 100\% & ~50.6 & 69.8 & 78.9~\\
			OccuSeg~\cite{occuseg}  & 100\% & ~48.6 & 67.2 & 78.8~\\
			PointGroup~\cite{pointgroup} & 100\% & ~40.7 & 63.6 & 77.8~\\
			3D-MPA~\cite{3dmpa} & 100\% & ~35.5 & 61.1 & 73.7~\\
			MTML~\cite{MTML} & 100\% & ~28.2 & 54.9 & 73.1~\\
			3D-BoNet~\cite{3DBoNet} & 100\% & ~25.3 & 48.8 & 68.7~\\
			3D-SIS~\cite{3DSIS} & 100\% & ~16.1 & 38.2 & 55.8~\\
			GSPN~\cite{gspn} & 100\% & ~15.8 & 30.6 & 54.4~\\
			\midrule
			One Obj One Pt Supervision: &&&&\\
			SegGroup (PointGroup) \cite{tao2020seggroup} & 0.02\% & ~24.6 & 44.5 & 63.7~ \\
            3D-WSIS \cite{tang20223dwsis} & 0.02\% & ~25.1 & 47.0 & 67.8~ \\
			\midrule
			\midrule
			\textbf{RWSeg (Ours)} & 0.02\% & ~34.8 & 56.7 & 73.9~ \\
			\bottomrule
		\end{tabular}
	\end{center}
	\vspace{-2mm}
	\caption{Instance segmentation results on ScanNet-v2 \cite{dai2017scannet} test set.}
	\label{ins_test}
\end{table}

\begin{table}[t]
	\footnotesize
	\begin{center}
    \scalebox{0.9}{
		\begin{tabular}{l|c|cc|cc}
			\toprule
            ~ & ~ & \multicolumn{2}{c}{\textbf{Area 5}} & \multicolumn{2}{c}{\textbf{6-fold}} \\
			Method & Supv. & mPre & mRec & mPre & mRec\\
			\midrule
			Full Supervision:~~ &&&&&\\
			PointGroup~\cite{pointgroup} & 100\% & 61.9 & 62.1 & 69.6 & 69.2\\
			\midrule
			One Obj One Pt Supervision: &&&&&\\
			SegGroup (PointGroup) \cite{tao2020seggroup} & 0.02\% & 47.2 & 34.9 & 56.7 & 43.3\\
            3D-WSIS \cite{tang20223dwsis} & 0.02\% & 50.8 & 38.9 & 59.3 & 46.7\\
			\midrule
			\midrule
			\textbf{RWSeg (Ours)} & 0.02\% & 60.1 & 45.8 & 68.9 & 56\\
			\bottomrule
		\end{tabular}
    }
	\end{center}
	\vspace{-2mm}
	\caption{Instance segmentation results on S3DIS \cite{S3DIS} dataset.}
	\label{ins_s3dis}
\end{table}

\subsection{Ablation Study}
In this section, we proceed to study the impacts of different components of our proposed method. Table \ref{tab:ablation_network} shows the network performance at different stages of training. We use ``Self-Attn" to represent the self-attention module in our network. In the setting of ``3D U-Net + Self-Attn", we freeze the backbone network and only train self-attention module, which shows the effectiveness of this component. Stage 1 training is supervised by initial weak labels. And Stage 2 training is supervised by the generated pseudo labels from our algorithm at the end of Stage 1. With our training strategy, the quality of semantic features can be steadily improved.

\begin{table} [h]
\centering
\scalebox{0.9}{
  \begin{tabular}{c|l|cc}
    \toprule
     mIoU & ~~~~~~~~Method & train set & val set \\
    \midrule
    Stage 1 & 3D U-Net  & 74.6 & 61.7  \\
    Stage 1 & 3D U-Net + Self-Attn  & 78.9 & 66 \\
    Stage 2 & 3D U-Net  & 80 & 68.4 \\
    Stage 2 & 3D U-Net + Self-Attn  & 81.6 &  70.3 \\
    \bottomrule
  \end{tabular}
}
\vspace{2mm}
\caption{Ablation study for network components. ``3D U-Net'' indicates our backbone network, and ``Self-Attn'' means our proposed self-attention module for feature propagation. Evaluated on ScanNet-v2 \cite{dai2017scannet} validation set.}
\label{tab:ablation_network}
\end{table}

\vspace{-6mm}

\paragraph{Ablations on Cross-graph Competing Random Walks (CRW)}
To make fair comparisons on clustering algorithms for pseudo label generation, we train a PointGroup \cite{pointgroup} backbone network with initial weak labels. On top of the shared network, we evaluate the performance of our CRW and other baseline methods in Table \ref{tab:ablation_CGCRW}. ``PointGroup BFS" represents a popular Breadth-First Search algorithm used in fully supervised 3D instance segmentation. K-means \cite{Hartigan1979} is a simple yet powerful unsupervised clustering algorithm to separate samples in $K$ groups of equal variance. Its character suits our task very well by nature. However, we found K-means is very sensitive to noise. The performance highly depends on the quality of semantic predictions and shift vectors. In contrast, our CRW is more robust and works well in different situations.

\begin{table} [h]
\centering
\scalebox{0.9}{
  \begin{tabular}{c|ccc}
    \toprule
     Baseline Methods & AP & AP{\tiny 50} & AP{\tiny 25} \\
    \midrule
    PointGroup BFS \cite{pointgroup} & 15.8  & 32.4 & 58.9  \\
    K-means$^\dag$ \cite{Hartigan1979} & 14.5  & 28.5 & 66.9 \\
    K-means$^\ddag$ \cite{Hartigan1979} & 23.5  & 44.1 & 72.5 \\
    \midrule
    \midrule
    \textbf{CRW$^\dag$ (Ours)} & 53.2  & 80.6 &  95.2 \\
    \textbf{CRW$^\ddag$ (Ours)} & \textbf{55}  & \textbf{82} &  \textbf{95.9} \\
    \bottomrule
  \end{tabular}
}
\vspace{2mm}
\caption{Comparison with pseudo label generation baseline methods on ScanNet-v2 \cite{dai2017scannet} training set. Methods marked with $^\dag$ are based on original coordinates. Methods marked with $^\ddag$ are based on shifted coordinates. BFS uses both sets of coordinates.}
\label{tab:ablation_CGCRW}
\end{table}

\begin{figure}[htb]
    \vspace{-2mm}
	\begin{center}
		\includegraphics[width=1.0\linewidth]{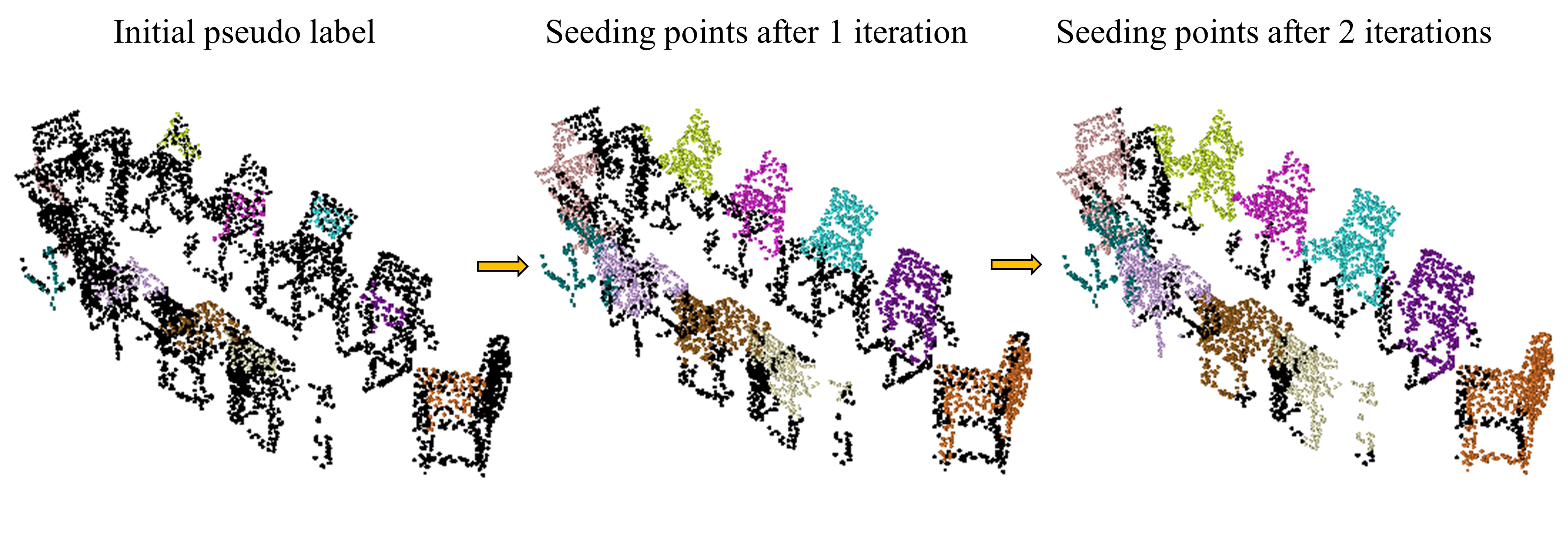}
	\end{center}
	\vspace{-4mm}
	\caption{Visualized example of CRW's seeding regions at different iterations.}
	\label{fig:CGCRW_ablation_iter}
\end{figure}

Figure \ref{fig:CGCRW_ablation_iter} shows the change of seeding regions during the process of Cross-graph Competing Random Walks. At each step, the top $50\%$ of the new predictions on unlabelled points are added as seed. It can be seen that new seeding points tend to be distributed at those regions relatively far from other seeds, as a result of cross-graph competition.

\begin{table} [h]
\centering
\scalebox{0.8}{
  \begin{tabular}{c|cccc}
    \toprule
     Iteration number (${t_2}_{max}$)  & 0 & 1 & 5 & 10 \\
    \midrule
    ~~~chair~~~~~~~~ AP $\uparrow$  & 64.2 & 66.3 & 67.4 & 67.4 \\
    bookshelf~~~ AP $\uparrow$  & 48.1 & 51.0 & 52.3 & 52.3 \\
    \bottomrule
  \end{tabular}
}

\vspace{2mm}
\caption{Impact of the competing mechanism and iteration number on CRW ($\theta=50\%$). Evaluated in AP for chair and bookshelf class on ScanNet-v2 \cite{dai2017scannet} training set. ${t_2}_{max}=0$ represents the converged results from the basic RWSeg without competing mechanism.}
\label{tab:ablation_CGCRW_iter}
\end{table}

The impact of using multiple random walk steps in Cross-graph Competing Random Walks (CRW) is shown in Table \ref{tab:ablation_CGCRW_iter}. As we expected, cross-graph competition is useful to resolve those ambiguous cases in instance segmentation, where objects from same category are compactly placed. Meanwhile, for those sparsely placed object categories, such as bathtub and door, their instance segments can already be well defined by proposed basic random walk algorithm. Competitions usually not exist for such cases. 

\section{Conclusion}
\label{sec:ccl}

In this paper, we propose a novel weakly supervised method for 3D instance segmentation on point clouds. With significantly fewer annotations, our network uses a self-attention module to propagate semantic features and a random walk based algorithm with cross-graph competition to generate high-quality pseudo labels. Comprehensive experiments show that our method achieves solid improvements on performance. The limitations of our method are discussed in the supplementary material.

\clearpage
\newpage

\section*{Acknowledgement}
This study is supported under the RIE2020 Industry Alignment Fund – Industry Collaboration Projects (IAF-ICP) Funding Initiative, as well as cash and in-kind contribution from the industry partner(s). This research work is also supported by the Agency for Science, Technology and Research (A*STAR) under its MTC Young Individual Research Grant (Grant No. M21K3c0130) and MTC Programmatic Funds (Grant No. M23L7b0021). This research is also partly supported by the MoE AcRF Tier 2 grant (MOE-T2EP20220-0007) and the MoE AcRF Tier 1 grant (RG14/22).

{\small
\bibliographystyle{ieee_fullname}
\bibliography{camera-ready} 

\begin{thebibliography}{10}\itemsep=-1pt

\bibitem{3d_weak_ins_box}
Point cloud instance segmentation with semi-supervised bounding-box mining.
\newblock {\em CoRR}, abs/2111.15210, 2021.

\bibitem{ahn2019weakly}
Jiwoon Ahn, Sunghyun Cho, and Suha Kwak.
\newblock Weakly supervised learning of instance segmentation with inter-pixel
  relations.
\newblock In {\em CVPR}, pages 2209--2218, 2019.

\bibitem{ahn2018learning}
Jiwoon Ahn and Suha Kwak.
\newblock Learning pixel-level semantic affinity with image-level supervision
  for weakly supervised semantic segmentation.
\newblock In {\em CVPR}, pages 4981--4990, 2018.

\bibitem{S3DIS}
Iro Armeni, Ozan Sener, Amir~R. Zamir, Helen Jiang, Ioannis Brilakis, Martin
  Fischer, and Silvio Savarese.
\newblock 3d semantic parsing of large-scale indoor spaces.
\newblock In {\em Proceedings of the IEEE International Conference on Computer
  Vision and Pattern Recognition}, 2016.

\bibitem{arun2020weakly}
Aditya Arun, CV Jawahar, and M~Pawan Kumar.
\newblock Weakly supervised instance segmentation by learning annotation
  consistent instances.
\newblock In {\em ECCV}, pages 254--270, 2020.

\bibitem{Chen_2021_ICCV}
Shaoyu Chen, Jiemin Fang, Qian Zhang, Wenyu Liu, and Xinggang Wang.
\newblock Hierarchical aggregation for 3d instance segmentation.
\newblock In {\em Proceedings of the IEEE/CVF International Conference on
  Computer Vision (ICCV)}, pages 15467--15476, October 2021.

\bibitem{Chen_HAIS_2021_ICCV}
Shaoyu Chen, Jiemin Fang, Qian Zhang, Wenyu Liu, and Xinggang Wang.
\newblock Hierarchical aggregation for 3d instance segmentation.
\newblock In {\em ICCV}, 2021.

\bibitem{chibane2021box2mask}
Julian Chibane, Francis Engelmann, Tuan~Anh Tran, and Gerard Pons-Moll.
\newblock Box2mask: Weakly supervised 3d semantic instance segmentation using
  bounding boxes.
\newblock In {\em European Conference on Computer Vision ({ECCV})}. {Springer},
  October 2022.

\bibitem{TWIST9879061}
Ruihang Chu, Xiaoqing Ye, Zhengzhe Liu, Xiao Tan, Xiaojuan Qi, Chi-Wing Fu, and
  Jiaya Jia.
\newblock Twist: Two-way inter-label self-training for semi-supervised 3d
  instance segmentation.
\newblock In {\em 2022 IEEE/CVF Conference on Computer Vision and Pattern
  Recognition (CVPR)}, pages 1090--1099, 2022.

\bibitem{dai2017scannet}
Angela Dai, Angel~X. Chang, Manolis Savva, Maciej Halber, Thomas Funkhouser,
  and Matthias Nie{\ss}ner.
\newblock Scannet: Richly-annotated 3d reconstructions of indoor scenes.
\newblock In {\em Proc. Computer Vision and Pattern Recognition (CVPR), IEEE},
  2017.

\bibitem{dong2023weakly}
Shichao Dong and Guosheng Lin.
\newblock Weakly supervised 3d instance segmentation without instance-level
  annotations, 2023.

\bibitem{dong2022learning}
Shichao Dong, Guosheng Lin, and Tzu-Yi Hung.
\newblock Learning regional purity for instance segmentation on 3d point
  clouds.
\newblock In {\em European Conference on Computer Vision}, pages 56--72.
  Springer, 2022.

\bibitem{9552005}
Nico Engel, Vasileios Belagiannis, and Klaus Dietmayer.
\newblock Point transformer.
\newblock {\em IEEE Access}, 9:134826--134840, 2021.

\bibitem{3dmpa}
Francis Engelmann, Martin Bokeloh, Alireza Fathi, Bastian Leibe, and Matthias
  Nießner.
\newblock 3d-mpa: Multi proposal aggregation for 3d semantic instance
  segmentation, 2020.

\bibitem{felzenszwalb2004efficient}
Pedro~F. Felzenszwalb and Daniel~P. Huttenlocher.
\newblock Efficient graph-based image segmentation.
\newblock {\em IJCV}, 59(2):167--181, 2004.

\bibitem{multiview2}
Zan Gojcic, Caifa Zhou, Jan~D. Wegner, Leonidas~J. Guibas, and Tolga Birdal.
\newblock Learning multiview 3d point cloud registration.
\newblock In {\em IEEE/CVF Conference on Computer Vision and Pattern
  Recognition (CVPR)}, June 2020.

\bibitem{Submanifold}
Benjamin Graham and Laurens van~der Maaten.
\newblock Submanifold sparse convolutional networks.
\newblock {\em CoRR}, abs/1706.01307, 2017.

\bibitem{Guo_2021}
Meng-Hao Guo, Jun-Xiong Cai, Zheng-Ning Liu, Tai-Jiang Mu, Ralph~R. Martin, and
  Shi-Min Hu.
\newblock Pct: Point cloud transformer.
\newblock {\em Computational Visual Media}, 7(2):187–199, Apr 2021.

\bibitem{occuseg}
Lei Han, Tian Zheng, Lan Xu, and Lu Fang.
\newblock Occuseg: Occupancy-aware 3d instance segmentation, 2020.

\bibitem{Hartigan1979}
J.~A. Hartigan and M.~A. Wong.
\newblock A k-means clustering algorithm.
\newblock {\em JSTOR: Applied Statistics}, 28(1):100--108, 1979.

\bibitem{hou2021exploring}
Ji Hou, Benjamin Graham, Matthias Nie{\ss}ner, and Saining Xie.
\newblock Exploring data-efficient 3d scene understanding with contrastive
  scene contexts.
\newblock In {\em Proceedings of the IEEE/CVF Conference on Computer Vision and
  Pattern Recognition}, pages 15587--15597, 2021.

\bibitem{huang2018weakly}
Zilong Huang, Xinggang Wang, Jiasi Wang, Wenyu Liu, and Jingdong Wang.
\newblock Weakly-supervised semantic segmentation network with deep seeded
  region growing.
\newblock In {\em CVPR}, pages 7014--7023, 2018.

\bibitem{MVPNet}
Maximilian Jaritz, Jia-Yuan Gu, and Hao Su.
\newblock Multi-view pointnet for 3d scene understanding.
\newblock {\em ArXiv}, abs/1909.13603, 2019.

\bibitem{3DSIS}
Hou Ji, Angela Dai, and Matthias Nie{\ss}ner.
\newblock 3d-sis: 3d semantic instance segmentation of rgb-d scans.
\newblock In {\em Proc. Computer Vision and Pattern Recognition (CVPR), IEEE},
  2019.

\bibitem{pointgroup}
Li Jiang, Hengshuang Zhao, Shaoshuai Shi, Shu Liu, Chi-Wing Fu, and Jiaya Jia.
\newblock Pointgroup: Dual-set point grouping for 3d instance segmentation,
  2020.

\bibitem{MTML}
Jean Lahoud, Bernard Ghanem, Marc Pollefeys, and Martin~R. Oswald.
\newblock 3d instance segmentation via multi-task metric learning, 2019.

\bibitem{multiview}
Lei Li, Siyu Zhu, Hongbo Fu, Ping Tan, and Chiew-Lan Tai.
\newblock End-to-end learning local multi-view descriptors for 3d point clouds.
\newblock In {\em IEEE/CVF Conference on Computer Vision and Pattern
  Recognition (CVPR)}, June 2020.

\bibitem{PointCNN}
Yangyan Li, Rui Bu, Mingchao Sun, Wei Wu, Xinhan Di, and Baoquan Chen.
\newblock Pointcnn: Convolution on x-transformed points.
\newblock In {\em NeurIPS}, pages 820--830. Curran Associates, Inc., 2018.

\bibitem{Liang_2021_ICCV}
Zhihao Liang, Zhihao Li, Songcen Xu, Mingkui Tan, and Kui Jia.
\newblock Instance segmentation in 3d scenes using semantic superpoint tree
  networks.
\newblock In {\em Proceedings of the IEEE/CVF International Conference on
  Computer Vision (ICCV)}, pages 2783--2792, October 2021.

\bibitem{SSTNet_liang2021instance}
Zhihao Liang, Zhihao Li, Songcen Xu, Mingkui Tan, and Kui Jia.
\newblock Instance segmentation in 3d scenes using semantic superpoint tree
  networks.
\newblock In {\em Proceedings of the IEEE/CVF International Conference on
  Computer Vision}, pages 2783--2792, 2021.

\bibitem{Lin2018Supervoxel}
Yangbin Lin, Cheng Wang, Dawei Zhai, Wei Li, and Jonathan Li.
\newblock Toward better boundary preserved supervoxel segmentation for 3d point
  clouds.
\newblock {\em ISPRS Journal of Photogrammetry and Remote Sensing}, 143:39 --
  47, 2018.
\newblock ISPRS Journal of Photogrammetry and Remote Sensing Theme Issue
  “Point Cloud Processing”.

\bibitem{MASC}
Chen Liu and Yasutaka Furukawa.
\newblock {MASC:} multi-scale affinity with sparse convolution for 3d instance
  segmentation.
\newblock {\em CoRR}, 2019.

\bibitem{otoc_Liu_2021_CVPR}
Zhengzhe Liu, Xiaojuan Qi, and Chi-Wing Fu.
\newblock One thing one click: A self-training approach for weakly supervised
  3d semantic segmentation.
\newblock In {\em Proceedings of the IEEE/CVF Conference on Computer Vision and
  Pattern Recognition (CVPR)}, pages 1726--1736, June 2021.

\bibitem{papandreou2015weakly}
George Papandreou, Liang-Chieh Chen, Kevin~P. Murphy, and Alan~L. Yuille.
\newblock Weakly-and semi-supervised learning of a deep convolutional network
  for semantic image segmentation.
\newblock In {\em ICCV}, pages 1742--1750, 2015.

\bibitem{JSIS3D}
Quang-Hieu Pham, Thanh Nguyen, Binh-Son Hua, Gemma Roig, and Sai-Kit Yeung.
\newblock Jsis3d: Joint semantic-instance segmentation of 3d point clouds with
  multi-task pointwise networks and multi-value conditional random fields.
\newblock In {\em The IEEE Conference on Computer Vision and Pattern
  Recognition (CVPR)}, 2019.

\bibitem{pinheiro2015image}
Pedro~O. Pinheiro and Ronan Collobert.
\newblock From image-level to pixel-level labeling with convolutional networks.
\newblock In {\em CVPR}, pages 1713--1721, 2015.

\bibitem{PointNet}
Charles~Ruizhongtai Qi, Hao Su, Kaichun Mo, and Leonidas~J. Guibas.
\newblock Pointnet: Deep learning on point sets for 3d classification and
  segmentation.
\newblock {\em 2017 IEEE Conference on Computer Vision and Pattern Recognition
  (CVPR)}, pages 77--85, 2016.

\bibitem{PointNet++}
Charles~Ruizhongtai Qi, Li Yi, Hao Su, and Leonidas~J. Guibas.
\newblock Pointnet++: Deep hierarchical feature learning on point sets in a
  metric space.
\newblock In {\em NIPS}, 2017.

\bibitem{tang20223dwsis}
Linghua Tang, Le Hui, and Jin Xie.
\newblock Learning inter-superpoint affinity for weakly supervised 3d instance
  segmentation.
\newblock In {\em ACCV}, 2022.

\bibitem{tao2020seggroup}
An Tao, Yueqi Duan, Yi Wei, Jiwen Lu, and Jie Zhou.
\newblock {SegGroup}: Seg-level supervision for {3D} instance and semantic
  segmentation.
\newblock {\em arXiv preprint}, 2020.

\bibitem{KPConv}
Hugues Thomas, Charles~R. Qi, Jean-Emmanuel Deschaud, Beatriz Marcotegui,
  François Goulette, and Leonidas~J. Guibas.
\newblock Kpconv: Flexible and deformable convolution for point clouds.
\newblock {\em ArXiv}, abs/1904.08889, 2019.

\bibitem{NIPS2017_3f5ee243}
Ashish Vaswani, Noam Shazeer, Niki Parmar, Jakob Uszkoreit, Llion Jones,
  Aidan~N Gomez, \L~ukasz Kaiser, and Illia Polosukhin.
\newblock Attention is all you need.
\newblock In I. Guyon, U.~V. Luxburg, S. Bengio, H. Wallach, R. Fergus, S.
  Vishwanathan, and R. Garnett, editors, {\em Advances in Neural Information
  Processing Systems}, volume~30. Curran Associates, Inc., 2017.

\bibitem{vu2022softgroup}
Thang Vu, Kookhoi Kim, Tung~M. Luu, Xuan~Thanh Nguyen, and Chang~D. Yoo.
\newblock Softgroup for 3d instance segmentation on 3d point clouds.
\newblock In {\em CVPR}, 2022.

\bibitem{wang2020weakly}
Haiyan Wang, Xuejian Rong, Liang Yang, Jinglun Feng, Jizhong Xiao, and Yingli
  Tian.
\newblock Weakly supervised semantic segmentation in 3{D} graph-structured
  point clouds of wild scenes.
\newblock {\em arXiv preprint arXiv:2004.12498}, 2020.

\bibitem{SGPN}
Weiyue Wang, Ronald Yu, Qiangui Huang, and Ulrich Neumann.
\newblock Sgpn: Similarity group proposal network for 3d point cloud instance
  segmentation.
\newblock In {\em CVPR}, 2018.

\bibitem{ASIS}
Xinlong Wang, Shu Liu, Xiaoyong Shen, Chunhua Shen, and Jiaya Jia.
\newblock Associatively segmenting instances and semantics in point clouds.
\newblock In {\em CVPR}, 2019.

\bibitem{Wei_2020_CVPR}
Jiacheng Wei, Guosheng Lin, Kim-Hui Yap, Tzu-Yi Hung, and Lihua Xie.
\newblock Multi-path region mining for weakly supervised 3d semantic
  segmentation on point clouds.
\newblock In {\em IEEE/CVF Conference on Computer Vision and Pattern
  Recognition (CVPR)}, June 2020.

\bibitem{wei2020multi}
Jiacheng Wei, Guosheng Lin, Kim-Hui Yap, Tzu-Yi Hung, and Lihua Xie.
\newblock Multi-path region mining for weakly supervised 3d semantic
  segmentation on point clouds.
\newblock In {\em Proceedings of the IEEE/CVF Conference on Computer Vision and
  Pattern Recognition}, pages 4384--4393, 2020.

\bibitem{PointConv}
Wenxuan Wu, Zhongang Qi, and Li Fuxin.
\newblock Pointconv: Deep convolutional networks on 3d point clouds.
\newblock {\em arXiv preprint arXiv:1811.07246}, 2018.

\bibitem{xu2020weakly}
Xun Xu and Gim~Hee Lee.
\newblock Weakly supervised semantic point cloud segmentation: Towards 10x
  fewer labels.
\newblock In {\em CVPR}, pages 13706--13715, 2020.

\bibitem{3DBoNet}
Bo Yang, Jianan Wang, Ronald Clark, Qingyong Hu, Sen Wang, Andrew Markham, and
  Niki Trigoni.
\newblock Learning object bounding boxes for 3d instance segmentation on point
  clouds, 2019.

\bibitem{gspn}
Li Yi, Wang Zhao, He Wang, Minhyuk Sung, and Leonidas Guibas.
\newblock Gspn: Generative shape proposal network for 3d instance segmentation
  in point cloud.
\newblock {\em arXiv preprint arXiv:1812.03320}, 2018.

\bibitem{PointWeb}
Hengshuang Zhao, Li Jiang, Chi-Wing Fu, and Jiaya Jia.
\newblock {PointWeb}: Enhancing local neighborhood features for point cloud
  processing.
\newblock In {\em CVPR}, 2019.

\bibitem{zhao2021pointtransformer}
Hengshuang Zhao, Li Jiang, Jiaya Jia, Philip Torr, and Vladlen Koltun.
\newblock Point transformer.
\newblock In {\em ICCV}, 2021.

\bibitem{zhou2018weakly}
Yanzhao Zhou, Yi Zhu, Qixiang Ye, Qiang Qiu, and Jianbin Jiao.
\newblock Weakly supervised instance segmentation using class peak response.
\newblock In {\em CVPR}, pages 2209--2218, 2018.

\end{thebibliography}


\begin{thebibliography}{10}\itemsep=-1pt

\bibitem{dai2017scannet}
Angela Dai, Angel~X. Chang, Manolis Savva, Maciej Halber, Thomas Funkhouser,
  and Matthias Nie{\ss}ner.
\newblock Scannet: Richly-annotated 3d reconstructions of indoor scenes.
\newblock In {\em Proc. Computer Vision and Pattern Recognition (CVPR), IEEE},
  2017.

\bibitem{Submanifold}
Benjamin Graham and Laurens van~der Maaten.
\newblock Submanifold sparse convolutional networks.
\newblock {\em CoRR}, abs/1706.01307, 2017.

\bibitem{3DSemanticSegmentationWithSubmanifoldSparseConvNet}
Benjamin Graham, Martin Engelcke and Laurens van~der Maaten.
\newblock 3D Semantic Segmentation with Submanifold Sparse Convolutional Networks.
\newblock {\em CoRR}, abs/1711.10275, 2017.

\bibitem{MASC}
Chen Liu and Yasutaka Furukawa.
\newblock {MASC:} multi-scale affinity with sparse convolution for 3d instance
  segmentation.
\newblock {\em CoRR}, 2019.

\bibitem{MTML}
Jean Lahoud, Bernard Ghanem, Marc Pollefeys, and Martin~R. Oswald.
\newblock 3d instance segmentation via multi-task metric learning, 2019.

\bibitem{pointgroup}
Li Jiang, Hengshuang Zhao, Shaoshuai Shi, Shu Liu, Chi-Wing Fu, and Jiaya Jia.
\newblock Pointgroup: Dual-set point grouping for 3d instance segmentation,
  2020.

\bibitem{occuseg}
Lei Han, Tian Zheng, Lan Xu, and Lu Fang.
\newblock Occuseg: Occupancy-aware 3d instance segmentation, 2020.

\bibitem{Chen_2021_ICCV}
Shaoyu Chen, Jiemin Fang, Qian Zhang, Wenyu Liu, and Xinggang Wang.
\newblock Hierarchical aggregation for 3d instance segmentation.
\newblock In {\em Proceedings of the IEEE/CVF International Conference on
  Computer Vision (ICCV)}, pages 15467--15476, October 2021.

\bibitem{Liang_2021_ICCV}
Zhihao Liang, Zhihao Li, Songcen Xu, Mingkui Tan, and Kui Jia.
\newblock Instance segmentation in 3d scenes using semantic superpoint tree
  networks.
\newblock In {\em Proceedings of the IEEE/CVF International Conference on
  Computer Vision (ICCV)}, pages 2783--2792, October 2021.

\bibitem{NIPS2017_3f5ee243}
Ashish Vaswani, Noam Shazeer, Niki Parmar, Jakob Uszkoreit, Llion Jones,
  Aidan~N Gomez, \L~ukasz Kaiser, and Illia Polosukhin.
\newblock Attention is all you need.
\newblock In I. Guyon, U.~V. Luxburg, S. Bengio, H. Wallach, R. Fergus, S.
  Vishwanathan, and R. Garnett, editors, {\em Advances in Neural Information
  Processing Systems}, volume~30. Curran Associates, Inc., 2017.

\bibitem{zhao2021pointtransformer}
Hengshuang Zhao, Li Jiang, Jiaya Jia, Philip Torr, and Vladlen Koltun.
\newblock Point transformer.
\newblock In {\em ICCV}, 2021.

\bibitem{S3DIS}
Iro Armeni, Ozan Sener, Amir~R. Zamir, Helen Jiang, Ioannis Brilakis, Martin
  Fischer, and Silvio Savarese.
\newblock 3d semantic parsing of large-scale indoor spaces.
\newblock In {\em Proceedings of the IEEE International Conference on Computer
  Vision and Pattern Recognition}, 2016.

\end{thebibliography}
}

\end{document}


\vspace{-20mm}
\title{Collaborative Propagation on Multiple Instance Graphs \\ for 3D Instance Segmentation with Single-point Supervision \\ (Supplementary Material)}
\vspace{-15mm}
\maketitle
    
\begin{figure*}
    \vspace{-10mm}
    \maketitle
	\begin{center}
		\includegraphics[width=1.0\linewidth]{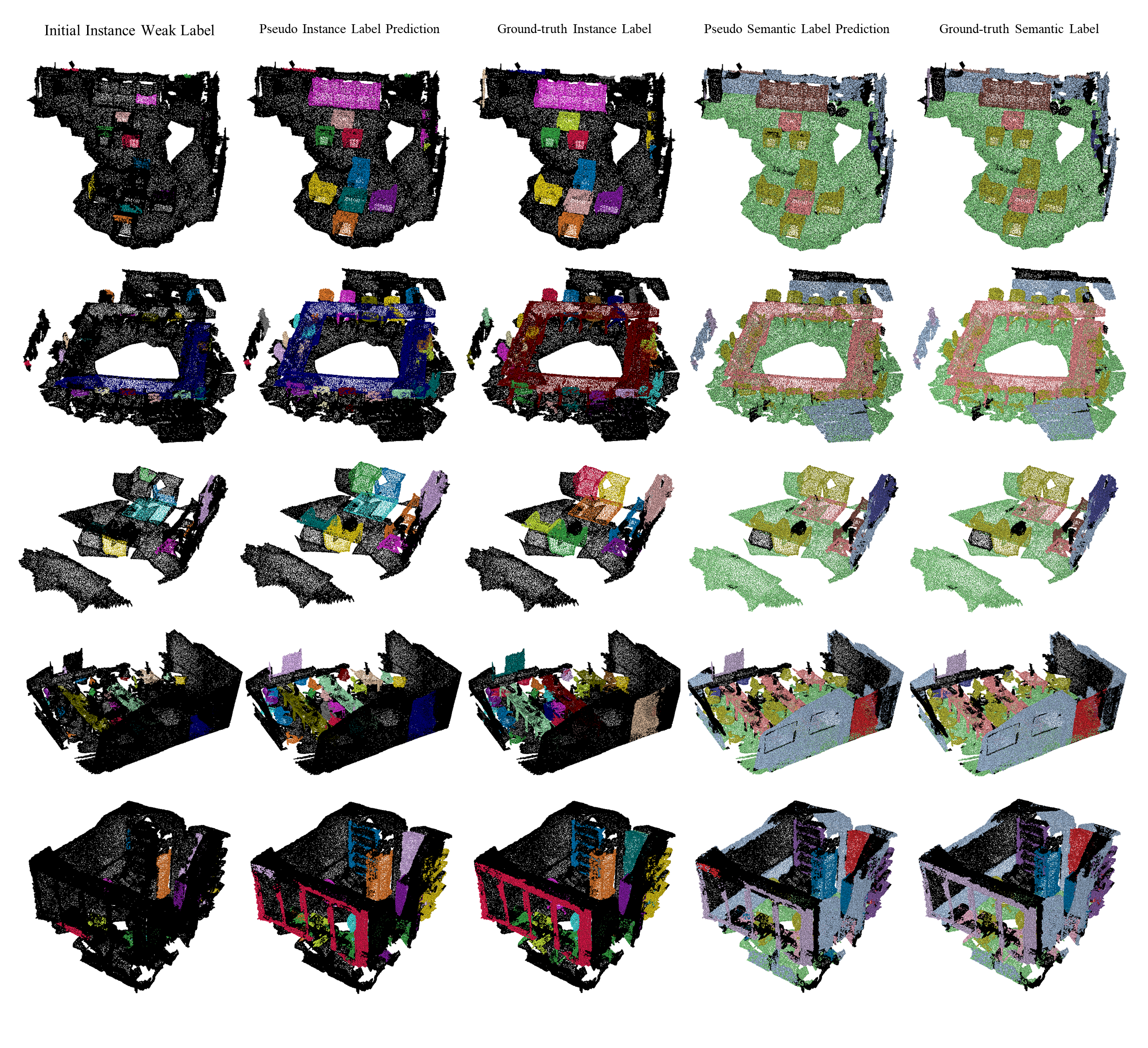}
	\end{center}
	\caption{More Qualitative Comparison on ScanNet v2 \cite{dai2017scannet} validation set.}
	\label{fig:scannet_result}
\end{figure*}
\vspace{20mm}

\section{Additional Qualitative Results}
In this section, we show some more visualization results of generated pseudo labels on the training set of ScanNet dataset \cite{dai2017scannet}. As shown in Figure \ref{fig:scannet_result}, initial instance weak labels are first derived from ``one object one point" weak annotations. Then, the proposed method RWSeg can propagate information to unlabelled points. Generated pseudo labels are compared with fully annotated ground-truth for semantic segmentation and instance segmentation respectively. The results show our high-quality pseudo labels have very similar patterns to the actual annotations and contain only minor errors.

\begin{figure*}[t]
	\begin{center}
		\includegraphics[width=0.99\linewidth]{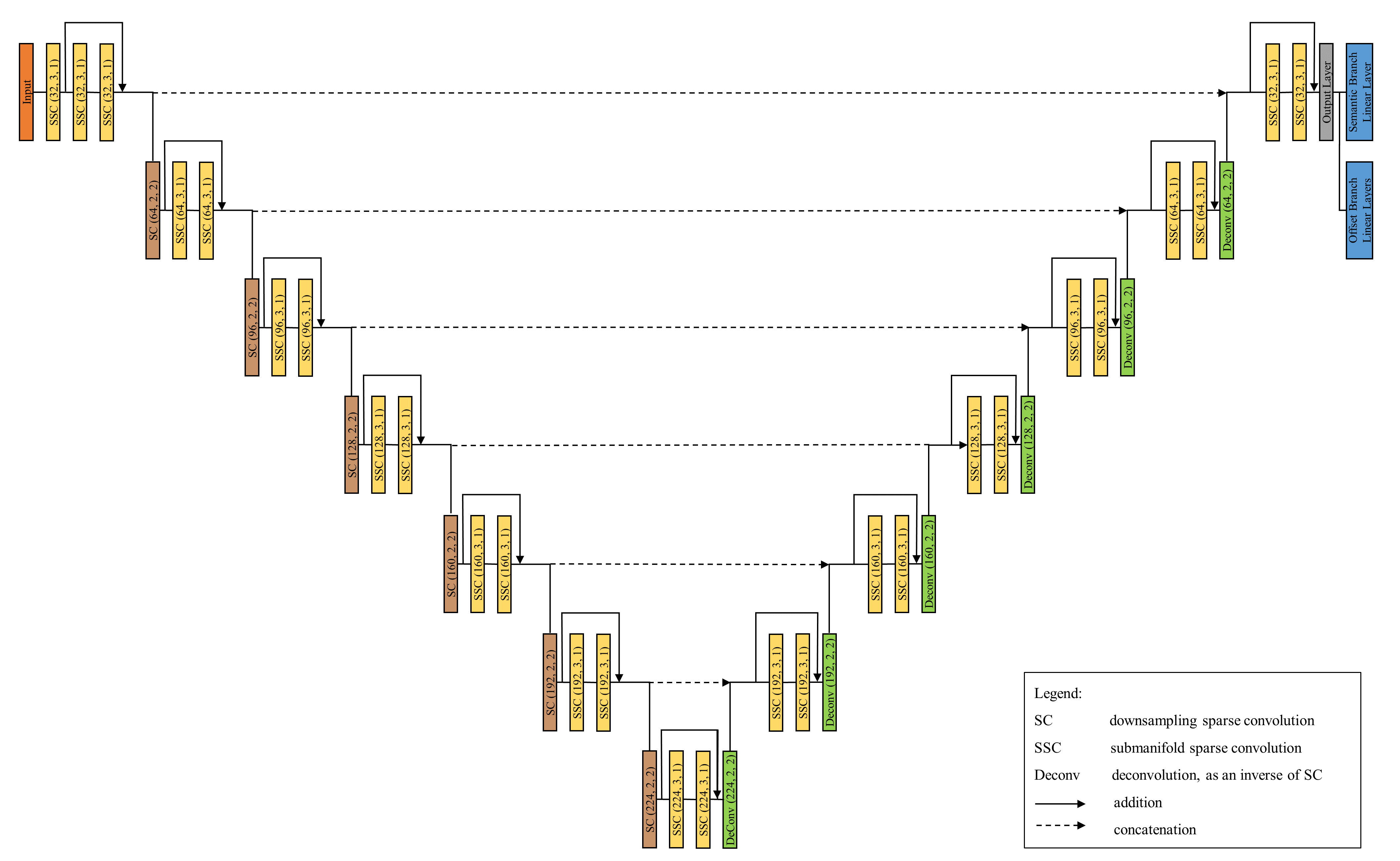}
	\end{center}
	\vspace{-4mm}
	\caption{The detailed structure of 3D U-Net backbone with submanifold sparse convolution \cite{Submanifold}.
    \vspace{-2mm}}
	\label{fig:backbone}
\end{figure*}

\section{Network Architecture Details}
In this section, we present the detailed structure of our 3D U-Net backbone with submanifold sparse convolution \cite{Submanifold} and self-attention module. The backbone network is originally introduced by Graham \cite{3DSemanticSegmentationWithSubmanifoldSparseConvNet} and has been widely used for feature extraction in point cloud segmentation tasks \cite{MASC,MTML,pointgroup,occuseg,Chen_2021_ICCV,Liang_2021_ICCV}. The core idea of submanifold sparse convolution is to efficiently process spatially-sparse data, otherwise using normal 3D convolution can be very computationally expensive.

\paragraph{Backbone network}
In Figure \ref{fig:backbone}, the backbone network takes the sparse voxelized representation of point cloud as input. The U-Net structure is mainly built based on sparse convolution (SC) layers and submanifold sparse convolution (SSC) layers. SC$(m, f, s)$ represents a downsampling sparse convolution (SC) layer with feature dimension $m$, kernel size $f$ and stride $s$. Residual connection is used to contain two submanifold sparse convolution (SSC) layers. Deconvolution represents an inverse operation of sparse convolution (SC). The output of the backbone network is split into the semantic branch and offset branch. The semantic branch further utilizes a self-attention layer for feature propagation.  For offset branch, point feature vectors are transformed via a two-layer MLP to the dimension of 3, which is then supervised by a regression loss for predicting the centroid shift vectors.

\begin{figure} [t]
	\begin{center}
		\includegraphics[width=0.36\linewidth]{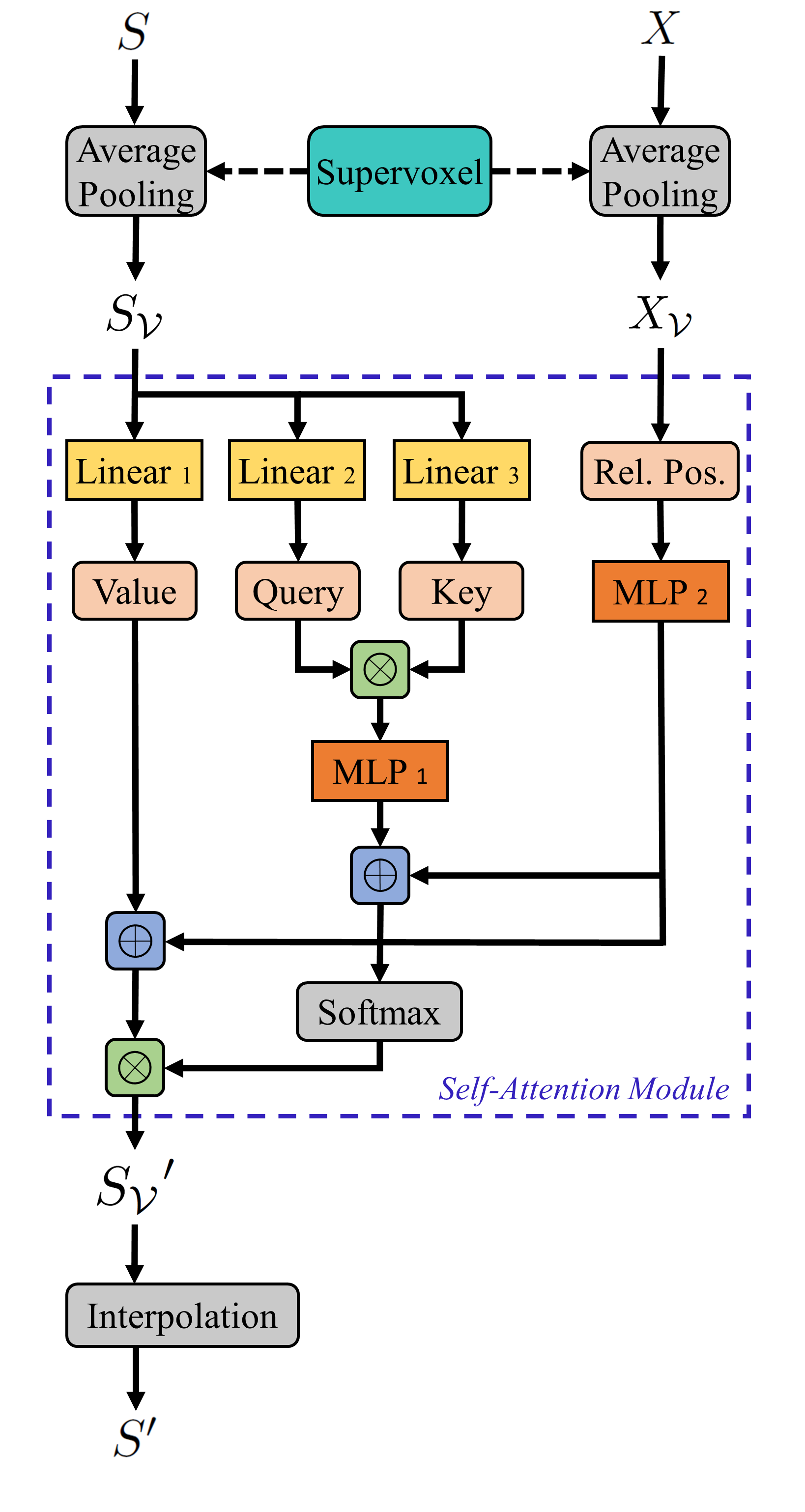}
	\end{center}
	\vspace{-8mm}
	\caption{Illustration of self-attention module in semantic branch for feature propagation. $\oplus$ denotes the broadcasting addition and $\otimes$ denotes the element-wise multiplication. Rel. Pos. represents the relative positional similarity of input coordinates.
    }
	\label{fig:attn_module}
\end{figure}

\paragraph{Self-attention module}
Figure \ref{fig:attn_module} illustrates the process of representing each supervoxel set $\mathcal{V}={p_1, p_2, ...,p_i}$ as a super-point. This is achieved by performing an average pooling operation on both the semantic features $S$ and the point coordinates $X$ for all points belonging to the set. Following \cite{NIPS2017_3f5ee243, zhao2021pointtransformer}, we first perform linear transformations of the input semantic features $S_{\mathcal{V}}$ to three matrices as query, key, and value ($Q, K, V$). Then, matrix $\bm A$ captures the similarity between queries and keys and also includes encoded positional information for adjustment. This can be written as

\begin{equation}\label{eqn:attn}
    \bm Q = S_{\mathcal{V}}W_Q,\quad \bm K = S_{\mathcal{V}}W_K,\quad \bm V = S_{\mathcal{V}}W_V,\\
\end{equation}

\begin{equation}\label{eqn:attn2}
    \bm A = \gamma (\frac{\bm Q \bm K^\top}{\sqrt{d}}) + \delta(X_{\mathcal{V}}, X_{\mathcal{V}}),
\end{equation}

where $d$ is the dimension of $Q$ and $V$, $X_{\mathcal{V}}$ the coordinates of supervoxels, $\gamma$ is a mapping function via MLP, $\delta$ is a positional similarity function via MLP. The output of self-attention can be formulated as

\begin{equation}\label{eqn:attn3}
    \mathrm{Attn}(S_{\mathcal{V}}) = \sigma({\bm A}) (\bar{\bm V} + \delta (X_{\mathcal{V}}, X_{\mathcal{V}})),
\end{equation}
where $\sigma(\cdot)$ is a softmax activation function. $\bar{\bm V}$ denotes a symmetric matrix created by repeatedly expanding $\bm V$. 

Lastly, refined semantic features are interpolated to the original size in point cloud. The training process is supervised by a conventional cross-entropy loss $H_{CE}$ with incomplete labels. We define the semantic loss as

\begin{equation}\label{eqn:ce}
	L_{sem} = - \frac{1}{N} \sum_{i=1}^{N} H_{CE}(y_i, \widehat{c}_i).
\end{equation}

where $\widehat{c}_i$ is the weak semantic label. Unlabelled points are ignored here. 

\paragraph{Offset loss function}
Following \cite{pointgroup}, We use a $L_{1}$ regression loss and a cosine similarity based direction loss to train the offset prediction,

\begin{equation}\label{eqn:offset_loss} 
    L_{offset} = \frac{1}{\sum_i m_i}\sum_i ||d_i - (\hat{q}_i - p_i)|| \cdot m_i - \frac{1}{\sum_i m_i} \sum_i \frac{d_i}{||d_i||_2} \cdot \frac{\hat{q}_i - p_i}{||\hat{q}_i - p_i||_2} \cdot m_i.
\end{equation}

where $\mathbf{m} = \{m_1, ..., m_N\}$ is a binary mask. The value of $m_i$ indicates whether point $i$ is on an instance or not. This means we only consider foreground points with weak labels for supervision.

\section{Ablations on Self-attention Module}
In Table \ref{tab:ablation_self_attn}, we perform ablation study on self-attention module by blocking relative position feature on ScanNet v2 \cite{dai2017scannet}. The structure with relative position feature broadcasting addition to both feature branch and attention branch can bring more performance gain.

\begin{table*} [htb]
	\vspace*{-0.5mm}
	\begin{center}
		\scalebox{1.0}[1.0]{
			\begin{tabular}{l | c| c }
			    \toprule[1pt]
				\small{Relative position usage} & \small{Train} & \small{Val}\\
				\hline
				\footnotesize{Baseline - backbone only} & \footnotesize{74.6} & \footnotesize{61.7} \\
				\footnotesize{None} & \footnotesize{77.3} & \footnotesize{64.1} \\
				\footnotesize{Feature branch only} & \footnotesize{77.6} & \footnotesize{64.3} \\
				\footnotesize{Attention branch only} & \footnotesize{78.3} & \footnotesize{65.3} \\
				\footnotesize{Feature branch + Attention branch} & \footnotesize{\textbf{78.9}} & \footnotesize{\textbf{66}}  \\
				\bottomrule[1pt]
			\end{tabular}
		}
	\end{center}
	\caption{Ablations on Self-attention Module}
	\label{tab:ablation_self_attn}
\end{table*}

\section{Random Walk with multiple Steps}
This section explains how to inference the equation as the final steady-state of the random walk algorithm (From Eq.6 to Eq.7 in original paper).

Random walk algorithm is performed by repeatedly adjusting node vector $b$ via transition matrix $\bm A$. At $t$-th iteration, the adjustment can be expressed as

\begin{equation}\label{eq:rw_1}
    {\bm b}^l_{t+1} = \alpha {\bm A} {\bm b}^l_{t} + (1-\alpha) {\bm b}^l_{0}, 
\end{equation}

where $b_{t}$ is the existing node vector derived at the previous random walk step, $b_0$ is the initial node vector, $\alpha \in [0, 1]$ is a blending coefficient between propagated score and initial score.

For random walk with multiple steps, we use $t$ to represent the $t$-th iteration and Expand Eq. (\ref{eq:rw_1}) to

\begin{equation}\label{eq:iter_rw_ex}
   {\bm b}^l_{t+1} =(\alpha \bm A)^{t+1} {\bm b}^l_{0} + (1 - \alpha)\sum_{i=0}^{t}(\alpha \bm A)^{i} {\bm b}^l_{0}.
\end{equation}

Applying $t \rightarrow  \infty$, since $\alpha \in [0,1]$, the first term in Eq. (\ref{eq:iter_rw_ex}) turns into
\begin{align}
	\underset{t\rightarrow\infty}{\lim}(\alpha \bm A)^{t+1}{\bm b}^l_0 = 0.
\end{align}
For the second term with matrices can be expanded as
\begin{equation}
   \underset{t\rightarrow \infty}{\lim} \sum_{i=0}^{t}(\alpha {\bm A})^{i} {\bm b}^l_0 = (\bm I - \alpha \bm A)^{-1} {\bm b}^l_0,
\end{equation}
where ${\bm I}$ is the identity matrix. Thus, the final steady-state of random walk algorithm can be written as
\begin{equation}
    {{\bm b}^l_{(\infty)}}   =  (1 - \alpha) ({\bm I} -\alpha {\bm A})^{-1} {\bm b}^l_0.
\end{equation}

\section{Competing Mechanism in CRW}
\begin{figure}[t]
	\begin{center}
		\includegraphics[width=0.9\linewidth]{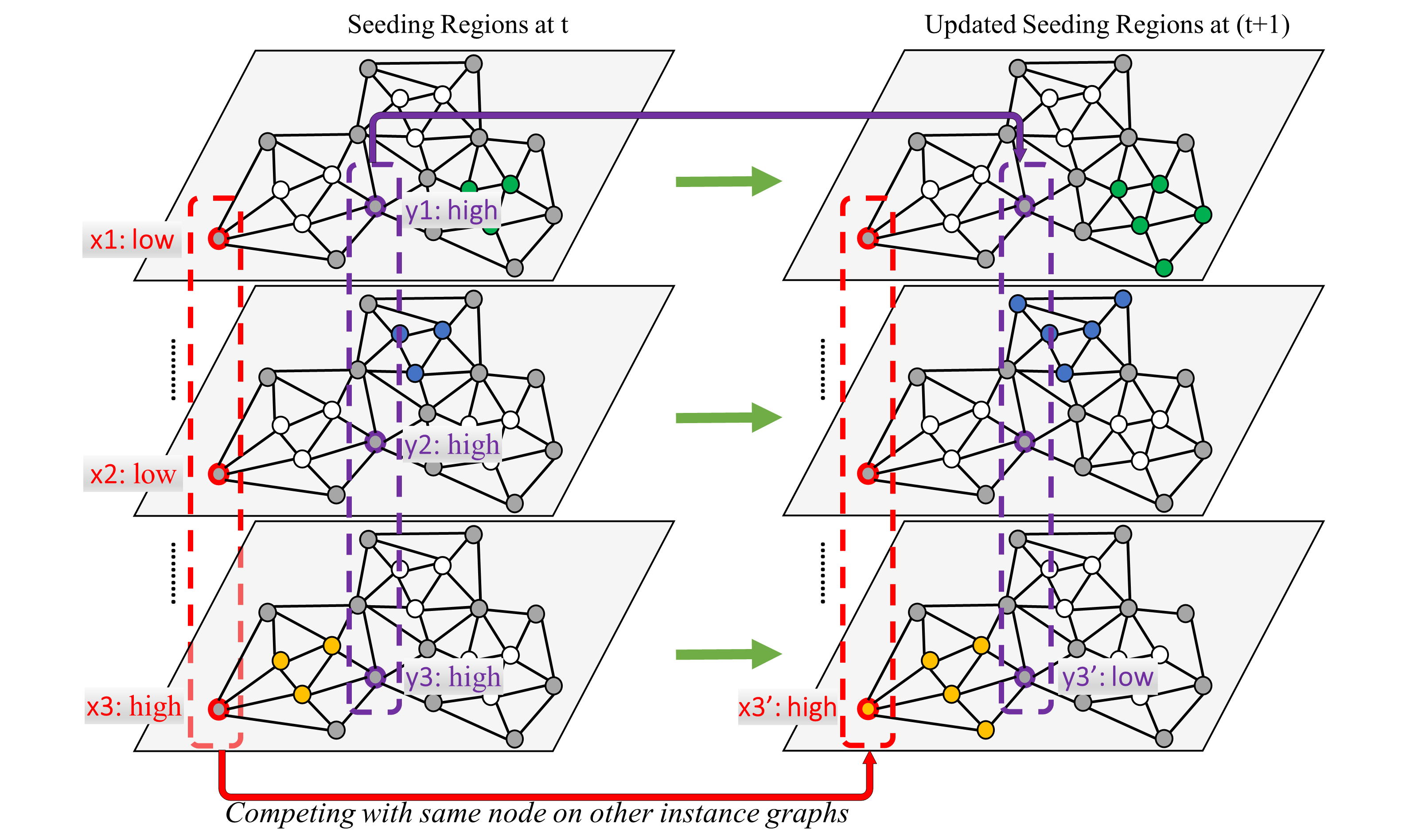}
	\end{center}
	\caption{Illustration of competing mechanism in CRW. This example shows the effect of competition on two different nodes (x3 in red, y3 in purple).
    \vspace{-1mm}}
	\label{fig:competing_mechanism}
\end{figure}

For illustrative purposes, we present an example in Figure \ref{fig:competing_mechanism}. In this case, the foreground category consists of three instance graphs, each with a distinct seeding point marked in green, blue, and yellow, respectively. Node $x3$ (in red) has two nodes in the same position, $x1$ and $x2$. At step $t$, their node scores are determined by their overall distance to the seeding points. Since $x1$ is far from the seeding points marked in green, its score will be low after a random walk step, whereas $x3$, which is closer to the seeding points marked in yellow, will have a higher score. After applying SoftMax normalization to the scores of $x1$, $x2$, and $x3$, the output score for $x3'$ at step $t+1$ will be high, as it faces less competition from the other two nodes.

Similarly, we have a node $y3$ with two same-positioned nodes, $y1$ and $y2$, placed in the center of three instances. At step $t$, the scores of $y1$, $y2$, and $y3$ are all high. However, during normalization, $y3$ receives a strong repulsive effect from $y1$ and $y2$. Thus, the output score $y3'$ at step $t+1$ will be low.

Finally, the proposed algorithm compares the node scores at step $t+1$. In this case, the node $x3'$ will have a higher priority to be grouped into seeds than $y3'$. This is because node $x3'$ is highly likely from the instance in yellow, whereas there is lower confidence in $y3'$. Therefore, we leave this node to be grouped in the later steps.

\section{Ablations on hyperparamters in CRW}
In Table \ref{tab:ablation_CGCRW_hyperparameter}, we show the experimental results with varying hyperparameters for the competing mechanism in CRW. The considered baseline is the proposed baseline random walk algorithm, which is represented by ${t_2}_{max}=0$. 

The table illustrates that a lower update percentage $\theta$ typically leads to better results but requires more iterations ${t_2}_{max}$, as it gradually groups the most confident points with our competing mechanism. The improvements over the random walk baseline are consistently observed. As discussed in the paper, the extent of the improvement depends on the distribution of the dataset. Notably, the proposed design in CRW is particularly effective in solving challenging cases, such as those with compacted objects of the same class.

\begin{table*}
	\vspace*{-0.5mm}
	\begin{center}
		\scalebox{1.0}[1.0]{
			\begin{tabular}{c | c | c | c }
			    \toprule[1pt]
				\small{Update percentage $\theta$} & \small{Iteration number ${t_2}_{max}$} & \small{AP (chair)} & \small{AP (bksf)} \\
				\hline
				N.A. & 0 & 64.2 &  48.1 \\
				80\% & 5 & 66.7 \textcolor{red}{(+2.5)} & 49.9 \textcolor{red}{(+1.8)} \\
				50\% & 5 & 67.4 \textcolor{red}{(+3.2)} & 52.3 \textcolor{red}{(+4.2)}\\
				20\% & 5 & 67 \textcolor{red}{(+2.8)} & 53.4 \textcolor{red}{(+5.3)}\\
				20\% & 20 & 67.3 \textcolor{red}{(+3.1)} & 54.4 \textcolor{red}{(+6.3)}\\
				10\% & 50 &  67.3 \textcolor{red}{(+3.1)} & 55.1 \textcolor{red}{(+7.0)}\\
				\bottomrule[1pt]
			\end{tabular}
	    }
	\end{center}
	\caption{Experiments with different CRW hyperparameters on ScanNet v2 \cite{dai2017scannet}}
	\label{tab:ablation_CGCRW_hyperparameter}
\end{table*}

\vspace{4mm}
\section{Additional Analysis in CRW}
After conducting experiments with various values of hyperparameters for ${t_1}{max}$ and $\alpha$, we have observed that our algorithm can converge after just a single random walk step. Further increasing the iteration number of ${t_1}{max}$ only results in marginal improvements. We argue that the fully connected graph used in our algorithm has a wide influence field. This means that it can exert its influence over a large area, and consequently, reduces the need for multiple random walk steps.

Hyperparameter $\alpha \in [0, 1]$  is used to control the trade-off between propagated node values and initial node values. Intuitively, it prevents deviating too fast from initial segmentation values. In our experiments, different values of $\alpha$ create a minor influence on the final converged results (less than $0.2\%$ in mAP). However, if we set the value of $\alpha$ to $1$ to remove the effect from initial values, the performance of random walk is dropped by $1\%$ in mAP.

\begin{figure}[t]
	\begin{center}
		\includegraphics[width=0.75\linewidth]{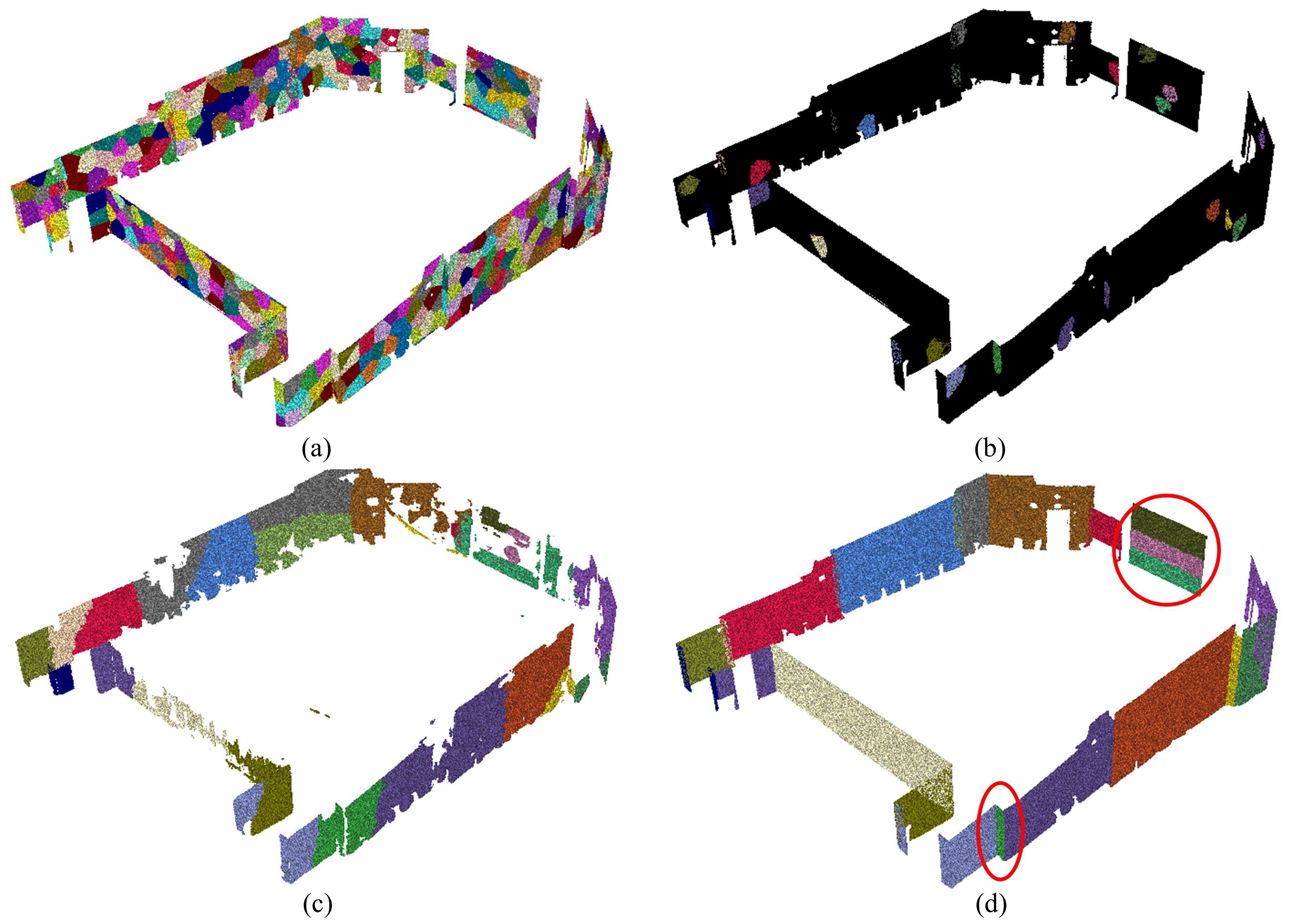}
	\end{center}
	\vspace{-4mm}
	\caption{Limitations of RWSeg on S3DIS \cite{S3DIS} dataset. (a) generated supervoxels (b) initial instance weak labels (c) generated instance pseudo labels (d) ground-truth instance labels
    \vspace{-1mm}}
	\label{fig:limitations}
\end{figure}

\section{Limitations of RWSeg}
In S3DIS \cite{S3DIS} dataset, some background stuff such as walls, ceilings, boards are also treated as instances by their setting. As shown in Figure \ref{fig:limitations} (d), walls are intentionally labeled as separate instances, even though they are part of the background. Additionally, these walls can vary greatly in size, which poses a challenge for our method. Our method is primarily designed for common instance types and may struggle to make accurate predictions on these cases, especially with limited initial weak labels. In practice, one possible solution is to use surface normals as a clue and apply unsupervised plane estimation methods. However, this is beyond the scope of this work and goes beyond our objectives.

\clearpage
\newpage

{\small
\bibliographystyle{ieee_fullname}
\bibliography{camera-ready_supp} 
}